\documentclass[lettersize,journal]{IEEEtran}
\usepackage{amsmath,amsfonts}
\usepackage{algorithm}
\usepackage{array}
\usepackage[caption=false,font=normalsize,labelfont=sf,textfont=sf]{subfig}
\usepackage{textcomp}
\usepackage{stfloats}
\usepackage{url}
\usepackage{verbatim}
\usepackage{graphicx}
\usepackage{cite}

\usepackage{times}
\usepackage{soul}
\usepackage{url}
\usepackage[utf8]{inputenc}
\usepackage[small]{caption}
\usepackage{amsthm}
\usepackage{booktabs}
\usepackage{algorithmicx}
\usepackage{algpseudocode}
\usepackage[switch]{lineno}
\usepackage{float}
\usepackage{subfig}
\usepackage{amssymb}
\usepackage{bm}
\usepackage{indentfirst}
\usepackage{color}
\usepackage{epsfig}
\usepackage{mathtools}
\usepackage{svg}
\usepackage{lineno}
\usepackage{bbm}
\usepackage[capitalize]{cleveref}

\usepackage{multirow}

\begin{document}

\title{Exploiting Minority Pseudo-Labels for Semi-Supervised Fine-grained Road Scene Understanding}

\author{Yuting Hong, Yongkang Wu, Hui Xiao, Huazheng Hao, Xiaojie Qiu, Baochen Yao and Chengbin Peng$^*$
	
	\thanks{Yuting Hong,  Yongkang Wu, Hui Xiao, Huazheng Hao, Baochen Yao and Chengbin Peng are with the Faculty of Electrical Engineering and Computer Science, Ningbo University, Ningbo, 315211, China.}
	\thanks{Xiaojie Qiu is with Zhejiang Cowain Automation Technology Co., Ltd,  Ningbo, 315200, China.}
	\thanks{$*$ Corresponding author: pengchengbin@nbu.edu.cn}
	}

\markboth{Journal of \LaTeX\ Class Files,~Vol.~14, No.~8, August~2021}%
{Shell \MakeLowercase{\textit{et al.}}: A Sample Article Using IEEEtran.cls for IEEE Journals}


\maketitle

\begin{abstract}

In fine-grained road scene understanding, semantic segmentation plays a crucial role in enabling vehicles to perceive and comprehend their surroundings. By assigning a specific class label to each pixel in an image, it allows for precise identification and localization of detailed road features, which is vital for high-quality scene understanding and downstream perception tasks. A key challenge in this domain lies in improving the recognition performance of minority classes while mitigating the dominance of majority classes, which is essential for achieving balanced and robust overall performance. 			However, traditional semi-supervised learning methods often train models overlooking the imbalance between classes. To address this issue, firstly, we propose a general training module that learns from all the pseudo-labels without a conventional filtering strategy. Secondly, we propose a professional training module to learn specifically from reliable minority-class pseudo-labels identified by a novel mismatch score metric.
The two modules are crossly supervised by each other so that it reduces model coupling which is essential for semi-supervised learning. During contrastive learning, to avoid the dominance of the majority classes in the feature space, we propose a strategy to assign evenly distributed anchors for different classes in the feature space. 
Experimental results on multiple public benchmarks show that our method surpasses traditional approaches in recognizing tail classes.
\end{abstract}


\section{Introduction}

Fine-grained road scene understanding is essential for intelligent transportation subsystems, including on-vehicle perception in autonomous cars, real-time traffic surveillance, and edge-based infrastructure monitoring, where accurate scene interpretation is critical to safety and performance. A fundamental technique to achieve this is semantic segmentation, which assigns a specific class label to each pixel in a road scene image, thereby identifying elements such as traffic signs, vehicles, pedestrians, cyclists, and other relevant obstacles \cite{ji2020encoder, chen2018encoder, xie2021segformer}. Given its role in enabling such detailed scene comprehension, a robust semantic segmentation method can significantly improve situational awareness and support accurate decision-making, making it indispensable for developing safe and efficient intelligent transportation systems.

Yet, supervised semantic segmentation for understanding large complex scenes is often costly due to the expensive annotation cost, and has sparked considerable interest in leveraging weakly labeled, easily labeled, and unlabeled data to boost model performance \cite{meng2019weakly, zhou2022context, zhao2022source, chen2021semi, chen2023adversarial, liao2024pda}. 
Among them,  semi-supervised semantic segmentation utilizes a small amount of labeled data with a large volume of unlabeled data to tackle this issue, 
generally revolve around three core architectures: mean teacher \cite{tarvainen2017mean, hu2021semi, wang2022semi}, cross confidence consistency \cite{ke2020guided}, and cross pseudo supervision \cite{chen2021semi}.In mean teacher architecture, the teacher model functions as an Exponential Moving Average (EMA) of the student model \cite{ke2019dual}. The other two architectures involve two student models with the same architecture learning from each other. Additionally, two primary learning strategies are employed in semi-supervised semantic segmentation: consistency regularization \cite{bachman2014learning,french2019semi,ouali2020semi} and entropy minimization \cite{arazo2020pseudo,lee2013pseudo,sohn2020fixmatch}. Consistency regularization guarantees that a classifier generates similar predictions for a sample when subjected to various perturbations, and entropy minimization is a self-training process where a model is supervised using pseudo-labels generated by another model.



\begin{figure}[t]
	\begin{center}
		\captionsetup[subfloat]{labelsep=none,format=plain,labelformat=empty}
		\subfloat{\includegraphics[width=1\linewidth]{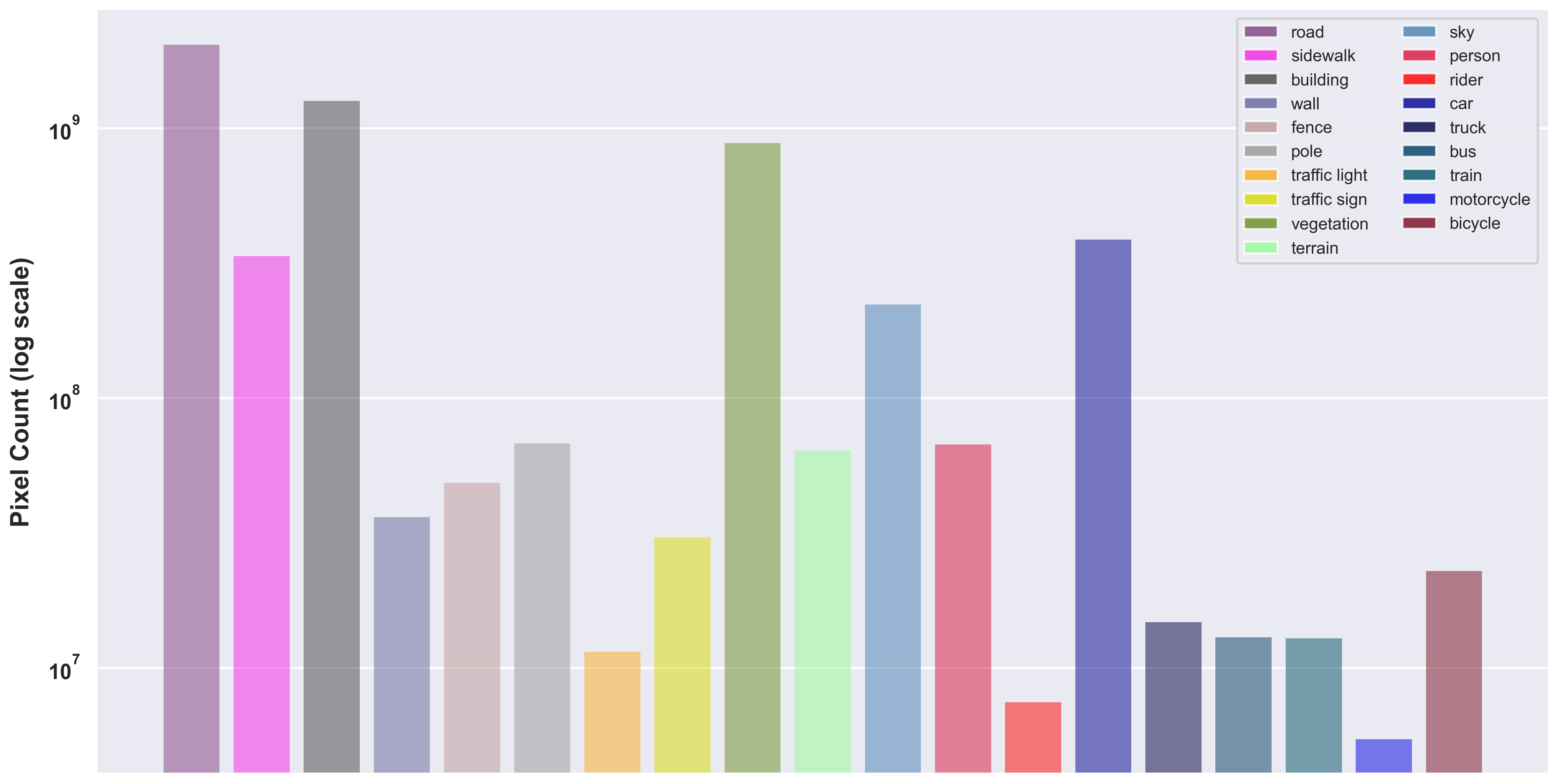}} \\
	\end{center}
	\vspace{-10pt}
	\caption{\textbf{Number of pixels in each class in Cityscapes dataset.} The Cityscapes dataset exhibits a significant class imbalance, as reflected in the distribution of finely annotated pixels across different classes. The number of pixels per class, represented on the y-axis, varies widely, with certain dominant classes occupying the majority of the dataset's pixel annotations, while other less prevalent classes account for only a small fraction. 
	}
	\label{distru}
	\vspace{-12pt}
\end{figure}

However, in the aforementioned semi-supervised semantic segmentation approaches, there are still  two critical challenges remaining. 
The first challenge lies in class imbalance, which is particularly pronounced in real-world road scene images. As shown in Fig. \ref{distru}, pixel classes often exhibit long-tailed distributions, where majority classes (e.g., roads, sky) dominate the dataset, while safety-critical minority classes (e.g., traffic lights, riders) are significantly underrepresented. Traditional methods \cite{cao2019learning,li2020overcoming,wei2021crest} are based on fixed prior knowledge for sampling or weighting, yet struggle to adapt to varying data characteristics\cite{hong2024multi}. To overcome this limitation, we introduce a cooperative training strategy consisting of a professional module that emphasizes minority-class pseudo-labels and a general module that learns from all pseudo-labels. This task-aware division improves flexibility while maintaining overall performance. In addition, to avoid the dominance of the majority classes in the feature space, we also introduce anchor contrastive learning such that different classes can be more evenly distributed in the embedded feature space by anchors.

The second challenge stems from model coupling. As training progresses, traditional mean-teacher models tend to become highly coupled \cite{ke2019dual,xiao2022semi}, causing them to learn highly similar semantic features. This limits their capacity to capture novel and useful information, ultimately hindering generalization. To alleviate this issue, we introduce a cross-training paradigm in which each student model is supervised by the teacher model from another module other than its own EMA \cite{tarvainen2017mean} counterpart. By applying differentiated learning strategies,  each module becomes semantically decoupled. 

In short, to cater to traffic scenarios,   we propose a \textbf{S}ynergistic \textbf{T}raining framework with \textbf{P}rofessional and \textbf{G}eneral Training (STPG) by synergistically integrating professional and general training modules with cross-guided decoupling mechanisms, which can address both class imbalance and model coupling issues, and the contributions are summarized as follows:

	\begin{itemize} 
		\item We propose a synergistic training framework with a professional module that leverages a mismatch score metric to identify and learn from minority class pseudo-labels, working alongside a general module with cross-supervision to reduce model coupling.

\item We propose an anchor contrastive learning strategy using evenly distributed anchors and optimal anchor-prototype matching to alleviate majority-class feature dominance and thus enhance class discrimination.

\item Our method significantly improves recognition of minority classes without sacrificing the performance of majority classes, leading to more balanced and robust segmentation.
	\end{itemize}
	
	\section{Related Work}
	
	\textbf{Semantic Segmentation and Road Scene Understanding} are essential for fine-grained perception of traffic environments, enabling accurate recognition of traffic signs, drivable areas, and other critical road elements, thereby forming the foundation for decision-making in intelligent transportation systems. Recent advancements have witnessed remarkable strides in semantic segmentation, which are largely propelled by deep 
	convolutional neural networks (CNNs)
	\cite{lecun2015deep} and transformers \cite{vaswani2017attention}. The leading-edge approaches in semantic segmentation \cite{cheng2022masked, guo2022segnext, zheng2021rethinking} predominantly adhere to the encoder-decoder paradigm \cite{long2015fully}. In this paradigm, the decoder replaces fully connected layers that are typically employed in classification tasks with convolutional layers, enabling pixel-wise predictions. For road scene understanding, SFNet-N\cite{muhammad2022vision} addresses boundary blurring in low-light images by integrating a light enhancement network with semantic information and a segmentation network with strong feature extraction capabilities. KGD\cite{wang2025knowledge} is a lightweight road segmentation framework for edge devices, where a student model learns from a teacher model using knowledge distillation to balance accuracy and efficiency. However, conventional supervised approaches heavily depend on large-scale, precisely annotated datasets, which can be prohibitively expensive to acquire in practice.

	\textbf{Semi-Supervised Learning} has seen significant advancements in recent years. Many of these methods share similar basic techniques, such as  consistency regularization \cite{duan2022mutexmatch, kim2022conmatch, chang2022fully,yin2025uncertainty} or pseudo-labeling \cite{zhou2023hypermatch, hu2021simple}. Consistency regularization is based on the clustering assumption that a decision boundary typically passes through a region with low sample density; thus, predictions are consistent under perturbations. Pseudo-labeling is based on the natural idea that predictions obtained from a model can be reused as supervision, which is motivated by entropy minimization. For example, FixMatch \cite{sohn2020fixmatch} is notable for its technique of generating pseudo-labels from weakly augmented unlabeled images and utilizing them to supervise models fed with strongly augmented images. Long-tail distribution has been extensively studied in semi-supervised learning. DST \cite{chen2022debiased} enhances the quality of pseudo-labels by adversarially optimizing representations to avoid worst-case scenarios. BaCon \cite{feng2024bacon} employs a meticulously crafted contrastive approach to directly regularize the distribution of instance representations.

	\textbf{Semi-Supervised Semantic Segmentation} aims to utilize both unlabeled data and labeled data to train a high-performance segmentation network that can assign pseudo-labels to images. Due to limited labeled data, the full utilization of unlabeled data has become an important issue. Preliminary works have focused on consistency regularization and entropy minimization. PRCL \cite{xie2023boosting} introduces a framework that enhances representation quality by incorporating its probability. MMFA \cite{yin2023semi} introduces a framework for feature augmentation that incorporates both multi-reliability and multi-level strategies to fully exploit pixel information. UniMatch\cite{yang2023revisiting} proposes a two-stream perturbation technique so that two strong views are simultaneously guided by a common weak view. CorrMatch \cite{sun2024corrmatch} introduces an adaptive threshold updating strategy and a pixel-wise similarity-based propagation of high-confidence predictions to efficiently leverage unlabeled data.
	However, the above semi-supervised semantic segmentation methods do not sufficiently explore the semantic information within minority classes. In this manuscript, we propose a synergistic training framework with two training modules and an anchor contrastive learning to resolve class imbalance issues.

	\textbf{Contrastive Learning} is widely used to learn similarity functions by pulling representations of similar samples closer and pushing dissimilar ones farther apart in the representation space. There are many different variations of contrastive learning for task-specific purposes. CLSA\cite{wang2022contrastive} proposes a general framework that leverages the distribution divergence between weakly and strongly augmented images to enhance retrieval and representation learning. PaCo\cite{cui2021parametric} proposes a parametric contrastive learning framework with class-wise learnable centers to address long-tailed recognition. DCL\cite{yeh2022decoupled} addresses the inefficiencies in traditional contrastive learning by removing the negative-positive-coupling effect. Recently, numerous contrastive learning methods \cite{alonso2021semi, zhong2021pixel, chen2022semi, ma2023enhanced} have been proposed for semi-supervised semantic segmentation.  PPS \cite{chen2022semi} employs automatic clustering and contrastive learning to manage outliers and establish clear decision boundaries. ESL\cite{ma2023enhanced} proposes a pixel-to-part contrastive learning method cooperated with an unsupervised object
	part grouping mechanism to improve its ability to distinguish between different classes. Different from previous contrastive learning approaches, in this work, aiming at addressing the class imbalance issue, we introduce  evenly distributed anchors in the feature space, which serve as reference points to improve class-level balance.
	%
	
	\section{Method}
	\begin{figure*}[t]
		
		\begin{center}
			
			\includegraphics[width=1.0\textwidth]{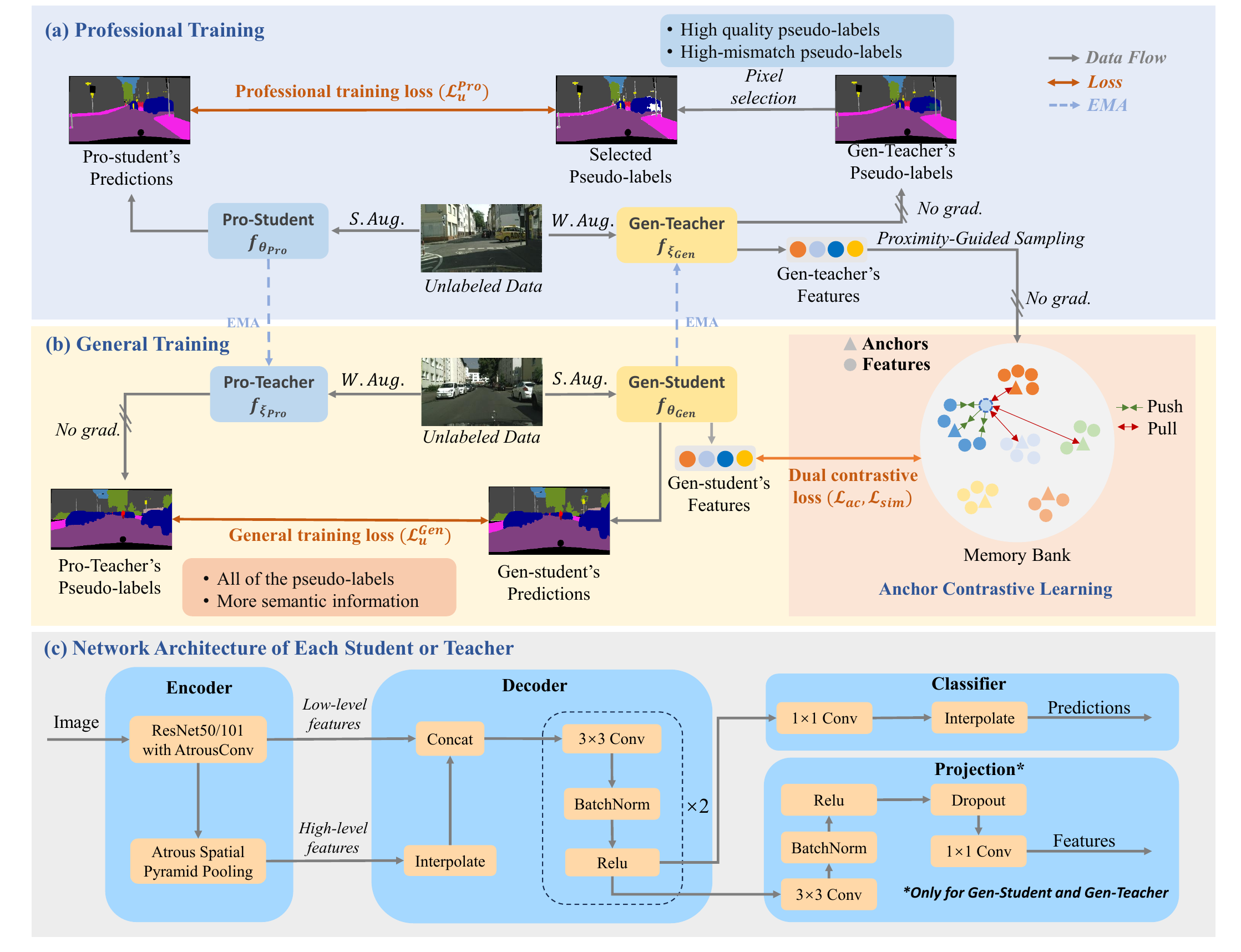}
			
		\end{center}
		\vspace{-13pt}
	\caption{\textbf{Overview of our framework.} There are two training parts: professional and general, as illustrated in parts (a) and (b).  In the professional training module, Pro-Student focusing on minority classes is supervised by selected pseudo-labels, and in the general training module,  Gen-Student is supervised by all pseudo-labels.
		Teachers are updated Exponential Moving Average (EMA) \cite{ke2019dual} to maintain a smoothed version of past student model parameters. 
		Anchor contrastive learning is designed  to foster more evenly distributed decision boundaries when training labels from different classes are imbalanced.  
		The supervised learning from labeled data is omitted for simplicity.  The network architecture of each student or teacher model is illustrated in part (c).
}
		\label{framework}
		\vspace{-12pt}
		
	\end{figure*}
	In this section, first, we define the problem. Second, we introduce the synergistic training framework. Last, we describe the utilization of anchor contrastive learning with anchors.
	
	\subsection{Problem Definition}
	\label{Preliminary}
	Let $\mathcal{D}$ be a dataset that consists of a labeled set $\mathcal{D}_l=\{(x_{l}^{i},y_{l}^{i})\}_{i=1}^{N_l}$ and an unlabeled set $\mathcal{D}_u=\{(x_{u}^{i})\}_{i=1}^{N_u}$, where $x^i$ is an input image and $y^i$ represents the corresponding pixel-level ground-truth. $N_l$ and $N_u$ denote the numbers of labeled images and unlabeled images, respectively, and generally ${N_l}\ll{N_u}$. In our approach, labeled data and unlabeled data are sampled equally in each training process. Our objective is to construct an effective semantic segmentation model by integrating a substantial quantity of unlabeled data with a small portion of labeled data.

	\subsection{Synergistic Training}
	\label{Dual-module Complementary Learning}
	\textbf{Overview.} To mitigate the accumulation of erroneous knowledge during training and reduce coupling, we employ a dual mean teacher architecture alongside mutual learning. In this setup, the parameters of the teacher model in one module are updated through EMA using the student model in other module. Fig. \ref{framework} summarizes the overall framework and the training procedure, which consists of two parallel student models, a general student model (Gen-Student)  $f_{\theta_{Gen}}$ and a professional student model (Pro-Student) $f_{\theta_{Pro}}$, each with different initializations. Their corresponding teacher models are general teacher model (Gen-Teacher) $f_{\xi_{Gen}}$ and professional teacher model (Pro-Teacher) $f_{\xi_{Pro}}$. Specifically, ${f}_{\theta_{Pro}}$ and ${f}_{\xi_{Gen}}$ form a professional training module, and ${f}_{\theta_{Gen}}$ and ${f}_{\xi_{Pro}}$ form a general training module, each focusing on a specific aspect. With the professional training module, Pro-Student and Pro-Teacher can acquire more information from high-quality samples or from minority classes. In contrast, we hope that the general training module can learn general feature representations from the whole dataset and the professional training module can focus on those from minority classes. Notably, all four models adhere to the same network architecture, which is based on DeepLabV3+\cite{chen2018encoder}. The specific architecture is illustrated in the Fig. \ref{framework}(c).
	
%
%
	
	\textbf{Supervised Loss.}
	For labeled images, as in most methods, a supervised loss is applied for Gen-Student and Pro-Student. Given the labeled image $x_l$ and its corresponding label $y_l$, the supervised loss $\mathcal{L}_s$ is written as:
	\begin{align}
		\mathcal{L}_{s}=\ell_{ce}({f}_{\theta_{Gen}}(\mathcal{A}^w(x_l)),y_l)+\ell_{ce}({f}_{\theta_{Pro}}(\mathcal{A}^w(x_l)),y_l),
	\end{align}
	where $\mathcal{A}^w(\cdot)$ denotes weak data augmentation and $\ell_{ce}$ is the pixel-wise cross-entropy loss function.
	
	%

	%
	%

	\textbf {Professional Training Module.}  
	The traditional learning approach tends to focus on majority classes and is not able to sufficiently learn minority classes. Thus, we propose a pixel selection strategy for producing refined pseudo-labels to improve the performance. The loss is calculated using high-quality and high-mismatch pseudo-labels from Gen-Teacher to supervise Pro-Student's predictions instead of all the pseudo-labels. The minority-class pseudo-labels are approximated with high-mismatch pseudo-labels, based on the observation that minority classes are more likely to be misclassified as other classes.
	
	For an unlabeled image, we obtain the prediction probabilities ${p}_{u}^{Pro} ,\hat{p}_{u}^{Gen} \in [0,1]^{W\times H\times C}$ from Pro-Student and Gen-Teacher and the pseudo-labels $\hat{y}_{u}^{Gen} \in \{0,1\}^{W\times H\times C}$ from Gen-Teacher:
	\begin{align}
		{p}_{u}^{Pro}=&{f}_{\theta_{Pro}}(\mathcal{A}^s(x_u)),\\
		\hat{p}_u^{Gen}=&{f}_{\xi_{Gen}}(\mathcal{A}^w(x_u)),\\
		\hat{y}_{u}^{Gen} =& OneHot(\hat{p}_u^{Gen}),
	\end{align}
	where $\mathcal{A}^s(\cdot)$ and $\mathcal{A}^w(\cdot)$ denote strong augmentation and weak data augmentation, respectively. The function $OneHot$ converts the probability into the one-hot format by assigning the most likely class for each pixel as one and the others as zero along the third axis of the matrix. Thus, only one entry along the third axis of each pixel in matrix $\hat{y}_{u}^{Gen}$ can be one.

	\begin{figure}[t]
		\begin{center}
			\captionsetup[subfloat]{labelsep=none,format=plain,labelformat=empty}
			\subfloat{\includegraphics[width=1.0\linewidth]{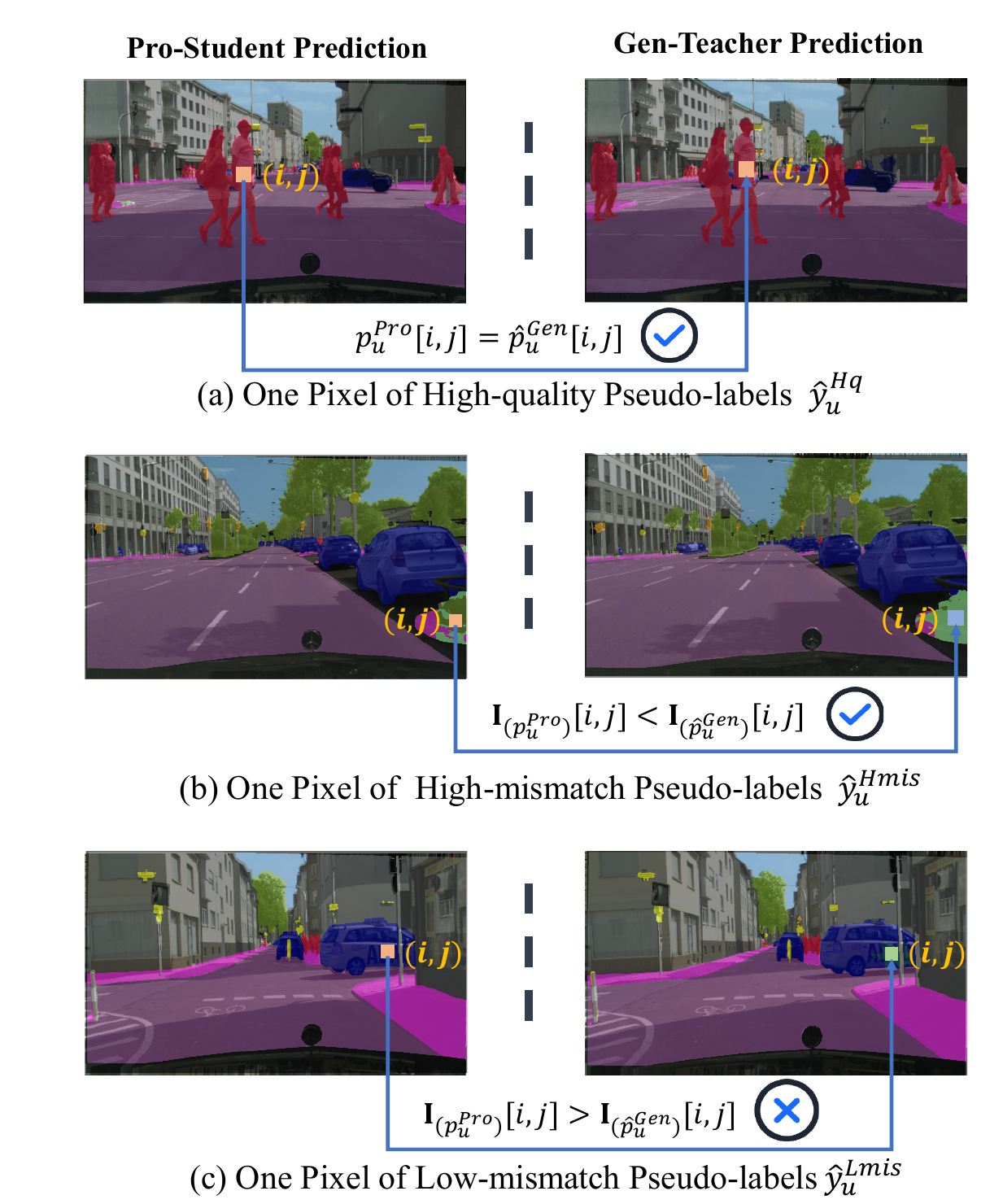}} \\
		\end{center}
		\vspace{-10pt}
		
		\caption{\textbf{Illustration of pixel selection strategy.} $\mathbf{I}(\cdot)[i,j]$ represents the predicted class for the pixel located at the coordinates $(i,j)$. (a) When the two predictions of a pixel are consistent, they are considered high-quality and used to train Pro-Student. (b) When the two predictions for a pixel are inconsistent, if the mismatch score of the class predicted by Gen-Teacher is larger than that of Pro-Student, the pixel usually contains more minority-class information and can be used for training Pro-Student. (c) 
			Otherwise, these pseudo-labels are not used for training Pro-Student.
		}
		\label{demo}
		\vspace{-12pt}
		
	\end{figure}
	
	In this work, the pseudo-labels represented by matrix $\hat{y}_{u}^{Gen}$ are divided into three mutually exclusive parts: high-quality pseudo-labels, high-mismatch pseudo-labels and low-mismatch pseudo-labels, represented by $ \hat{y}_{u}^{Hq}$, $\hat{y}_{u}^{Hmis}$, $\hat{y}_{u}^{Lmis} \in \{0,1\}^{W\times H\times C}$. For each pixel along the third axis, only one of the three matrices can have a nonzero entry; namely, we can apply the following restriction: $\hat{y}_{u}^{Hq} + \hat{y}_{u}^{Hmis} + \hat{y}_{u}^{Lmis}=\hat{y}_{u}^{Gen} $. 
	
	(1) High-quality pseudo-labels $\hat{y}_{u}^{Hq}$, as shown in Fig. \ref{demo}(a), are defined as the pseudo-labels that both Gen-Teacher and Pro-Student have consistently predicted. Inspired by DMT \cite{feng2022dmt}, we consider pseudo-labels with consistent pixel predictions across different training modules to be of high-quality. This is because the likelihood of both modules making the same error simultaneously is relatively low.
	
	(2) To identity mismatch pseudo-labels, we introduce a class-level mismatch score, which represents the proportion of a class being misclassified into other classes, based on the predictions of the Gen-teacher and Pro-student. This score reflects the degree of mismatch for a specific class. The high-mismatch pseudo-labels represented by $\hat{y}_{u}^{Hmis}$, as shown in Fig. \ref{demo}(b), are computed in three steps. 
	
\begin{figure}[t]
		\begin{center}
			\captionsetup[subfloat]{labelsep=none,format=plain,labelformat=empty}
			\subfloat{\includegraphics[width=0.65\linewidth]{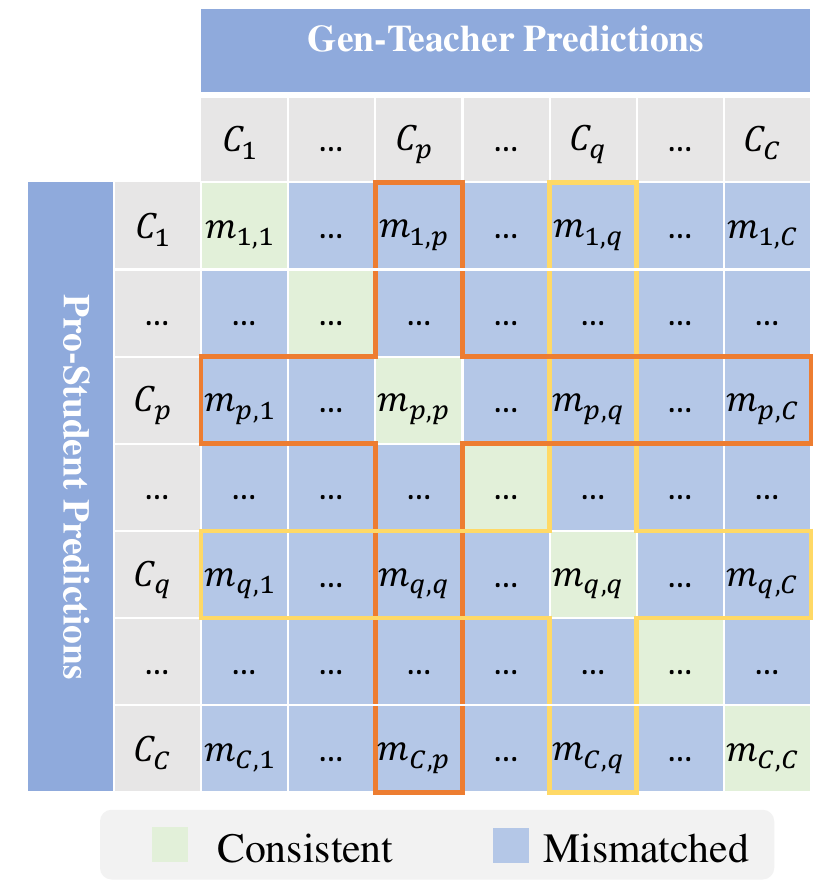}} \\
		\end{center}
		\vspace{-10pt}
		
		\caption{\textbf{Illustration of confusion matrix.} For each batch, we compute the confusion matrix between the predictions of Gen-Teacher and those of Pro-Student to obtain a mismatch score for each class. For example, $m_{p,q}$ is the number of pixels where Pro-Student's prediction is Class $p$ and Gen-Teacher's prediction is Class $q$ ($p,q \in [1,2,3,...,C]$). The proportions of mismatched predictions in the orange boxes and yellow boxes indicate the mismatch scores for Class $p$ and Class $q$, respectively.
		}
		\label{matrix}
		\vspace{-12pt}
		
	\end{figure}

	(i) We construct the confusion matrix $\mathbf{M}\in{\mathbb{R}^{C\times{C}}}$ with $C$ as the number of classes for the predictions of Pro-Student and Gen-Teacher 
	in each mini-batch. An illustration is shown in Fig. \ref{matrix}. 
	
	(ii) We define a class-level mismatch score vector $I$ by the confusion matrix, in which each element represents the proportion of mismatched predictions of Pro-Student and Gen-Teacher for a certain class. For example, we compute the mismatch score of class $q$ as follows:
	\begin{align}
		I_{q}=\frac{ {\textstyle \sum_{k=1}^{C}}m_{q,k}-m_{q,q}}{ {\textstyle \sum_{k=1}^{C}m_{q,k}} }  + \frac{ {\textstyle \sum_{k=1}^{C}}m_{k,q}-m_{q,q}}{ {\textstyle \sum_{k=1}^{C}m_{k,q}} },
	\end{align}
	where $m_{q,k}$ is the $(q,k)$-th entry of $\mathbf{M}$.  Here, the row term ${ {\textstyle (\sum_{k=1}^{C}}m_{q,k}-m_{q,q})}/{ {\textstyle \sum_{k=1}^{C}m_{q,k}}}$  measures the proportion of pixels that Pro-Student predicts as class $q$ but Gen-Teacher disagrees with. The column term ${ {\textstyle (\sum_{k=1}^{C}}m_{k,q}-m_{q,q})}/{ {\textstyle \sum_{k=1}^{C}m_{k,q}}}$ quantifies the proportion of pixels that Gen-Teacher predicts as class $q$ but Pro-Student disagrees with. 
		Thus, this combination offers a comprehensive measure of class-level disagreement: a high score reflects strong inconsistency between the two modules for Class $q$, which commonly occurs for minority or difficult classes. We therefore prioritize these classes, as they represent more challenging cases that require focused attention.
	
	We map the class-level mismatch scores onto the pixel predictions of both the Pro-Student and Gen-Teacher to generate two mismatch matrices, denoted as $\mathbf{I}_{(p_u^{Pro})}$ and $\mathbf{I}_{(\hat{p}_u^{Gen})}$. These matrices match the size of the input image, where each element corresponds to the mismatch score of the predicted class:
	\begin{equation}
		\mathbf{I}_{(\cdot)}[i,j] = I_{\arg\max_{k}(\cdot)[i,j,k]},
	\end{equation}
	Here, $I_{\arg\max_k(\cdot)[i,j,k]}$ is mismatch score associated with the predicted class  for the pixel at coordinates $(i,j)$.

	(iii) For inconsistent predictions between Pro-Student and Gen-Teacher, we select certain pseudo-labels to form high-mismatch pseudo-labels.
We first compute a binary mask matrix $\mathcal{M}_{Hmis}$, where each element is defined as:
		\begin{align}
		\mathcal{M}_{Hmis}[i,j]= \mathbb{I}\left[\mathbf{I}_{(p_u^{Pro})}[i,j] < \mathbf{I}_{(\hat{p}_u^{Gen})}[i,j]\right],
	\end{align}	
where $\mathbb{I}$ denotes indicator function. The inequality is performed element-wise, meaning that each element of $\mathcal{M}_{Hmis}$ is set to $1$ if the mismatch score of the Gen-Teacher's predicted class is greater than that of the Pro-Student's one for the corresponding pixel, and $0$ otherwise.

Using this mask, we generate the high-mismatch pseudo-labels $\hat{y}^{Hmis}_{u}$ by retaining only the valid pixels from the Gen-Teacher's predictions. Specifically, a pixel is considered valid if the corresponding element in the mask $\mathcal{M}_{Hmis}$ is 1, and invalid if the corresponding element is 0:
	\begin{align}
		\hat{y}^{Hmis}_{u} = \hat{y}^{Gen}_{u} \odot \mathcal{M}_{Hmis}.
	\end{align}	
	
	For example, for a certain pixel, if the Pro-Student's prediction is Class $p$ and the Gen-Teacher's prediction is Class $q$, and if $I_p < I_q$, we consider the Gen-Teacher's prediction to be a high-mismatch pseudo-label. These pseudo-labels have high mismatch scores, indicating that they are easily misled by other majority classes. As a result, the model fails to learn these classes effectively. Therefore, we encourage the model to fully explore the semantic information of these classes to enhance their learning.

	(3) The low-mismatch pseudo-labels $\hat{y}^{Lmis}_{u}$, as shown in Fig. \ref{demo}(c), are the remaining pseudo-labels that do not belong to $\hat{y}^{Hq}_{u}$ or $\hat{y}^{Lmis}_{u}$. These pseudo-labels are not used for training Pro-Student. 

	The professional training loss $\mathcal{L}_{u}^{Pro}$ is computed as follows:
	\begin{align}
		\mathcal{L}_{u}^{Pro} =\omega_u^{Pro} \ell_{ce}(p_u^{Pro}, \hat{y}_{u}^{Hq} + \hat{y}_{u}^{Hmis}), 
		\label{proce}
	\end{align} 
	where $\omega_u^{Pro}$ is a weighting matrix and each element of $\omega_u^{Pro}$ represents the confidence of a pseudo-label generated by the teacher model, i.e., 
	$\omega_u^{Pro}[i,j] = \max_{c \in \{1,...,C \}} \hat{p}_u ^{Gen}[i,j,c]$. For pseudo-labels not belonging to $\hat{y}_{u}^{Hq}$ or $\hat{y}_{u}^{Hmis}$, the corresponding elements in $\omega_u^{Pro}$ are zero.
	Using the prediction confidence of the teacher model as the weight of the student model to calculate the loss can reduce the impact of suspicious 
noisy labels. 
	Therefore, unlike traditional confidence thresholding approaches for filtering out numerous pseudo-labels, we can take advantage of unlabeled data.

Specifically, the high-quality pseudo-labels, high-mismatch pseudo-labels, and low-mismatch pseudo-labels generation rely only on simple matrix operations such as comparison, indexing, and element-wise addition. The time complexity for computing the confusion matrix in Step (i) involves $\mathcal{O}(N)$ operations, as each pixel is processed to update the matrix. Similarly, the mismatch score calculation in Step (ii) requires $\mathcal{O}(C^2)$ operations, as it involves summing over rows and columns for each class. However, the operations for matching matrices and corresponding masks in Step (iii) are dominated by $\mathcal{O}(N)$. Given that $N \gg C$, the overall time and space complexity remains linear with respect to the number of pseudo-labels, i.e., the number of pixels.

	\textbf{General Training Module.}
	For the general training module, we choose all pseudo-labels from Pro-Teacher as supervision for Gen-Student. For an unlabeled image, we obtain the prediction probability ${p}_{u}^{Gen}, \hat{p}_u^{Pro} \in [0,1]^{W\times H\times C}$ from Gen-Student and Pro-Teacher and the pseudo-labels $\hat{y}^{Pro}_u \in \{0,1\}^{W\times H\times C}$ from Pro-Teacher.
	\begin{align}
		{p}_{u}^{Gen}=&{f}_{\theta_{Gen}}(\mathcal{A}^s(x_u)),\\
		\hat{p}_u^{Pro} = &{f}_{\xi_{Pro}}(\mathcal{A}^w(x_u)),\\
		\hat{y}_{u}^{Pro} =& OneHot(\hat{p}_u^{Pro}),
	\end{align}
	The general training loss $\mathcal{L}_{u}^{Gen}$ is calculated as follows:
	\begin{align}
		\mathcal{L}_{u}^{Gen} = \omega_u^{Gen} \ell_{ce}(p_u^{Gen},\hat{y}^{Pro}_{u}),
	\end{align}
	We similarly used the reweighting method mentioned above, and $\omega[i,j] = \max_{c \in \{1,...,C \}} \hat{p}_u ^{Pro}[i,j,c]$

	
	\subsection{Anchor Contrastive Learning}
	\label{contrastive learning}
\begin{figure*}[t]
		\begin{center}
			\captionsetup[subfloat]{labelsep=none,format=plain,labelformat=empty}
			\subfloat{\includegraphics[width=1\linewidth]{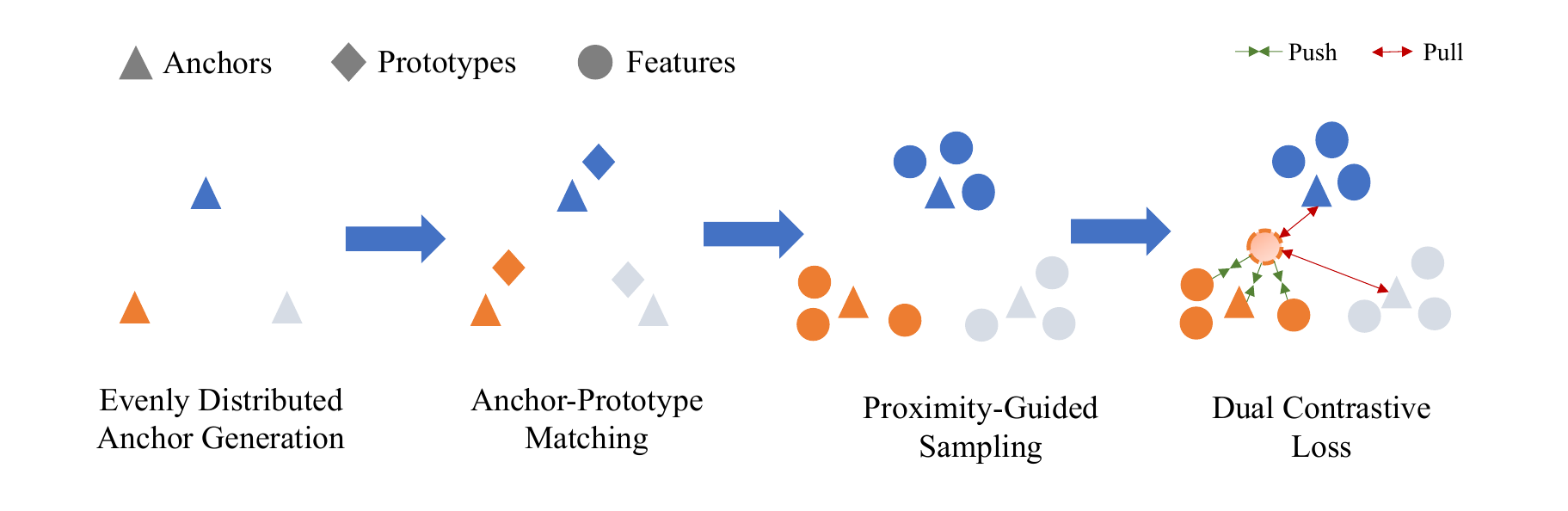}} \\	
		\end{center}
		\vspace{-15pt}
		
		\caption{\textbf{Illustration of anchor contrastive learning.} We generate anchors that are evenly distributed in the feature space and perform one-to-one matching with class prototypes. During training, we sample a large number of features near the anchors and combine them with the anchors for contrastive learning.
		}
		\label{acl}
		\vspace{-8pt}
		
	\end{figure*}

	Real-world datasets frequently display long-tail distributions, leading to significant class imbalances where dominant classes can distort model training and influence decision boundaries for minority classes. To address this issue, as shown in Fig. \ref{acl}, we propose an anchor contrastive learning strategy using evenly distributed anchors and optimal anchor-prototype matching to alleviate majority-class feature dominance and thus enhance class discrimination. This strategy aims to maintain a uniform distribution in the feature space for all classes, including minority classes. By doing so, it enhances decision boundaries and promotes better generalization, which is particularly effective for handling long-tailed data distributions.
	
\textbf{Evenly Distributed Anchors Generation.} Before training, we define $C$ as the number of classes and $v_i$ with $i = 1,2,3,...,C$ as anchors. The dimension of each anchor is equal to the dimension of the features, denoted as $D$, i.e., $v_i \in \mathbb{R}^D$.  These anchors are randomly initialized. To avoid any potential bias for initial values and ensure that they are in a comparable scale with the features, they are sampled from a standard normal distribution, i.e., $v_i \sim \mathcal{N}(0, I_D)$, and then normalized to have unit L2-norm.

To achieve an evenly distributed initial state, after random initialization, we apply an optimization step to ensure that each anchor maintains a similar distance from others. Let $V = \{v_1, v_2, \dots, v_C\}$ be the set of $C$ anchor vectors. We then perform the following objective minimization approach, adapted from \cite{wang2020understanding}:
\begin{equation}
	\min_{V} \left( \frac{1}{C}\sum_{i=1}^{C}\log\sum_{j=1}^{C}e^{v_i^T\cdot v_j/\tau } \right),
\end{equation}
where $T$ is matrix transpose. This objective function, designed with a temperature coefficient $\tau$, works to push anchors apart from one another, thereby ensuring they adopt a broadly spherical arrangement on the unit hypersphere.

	\textbf{Prototypes Update.} In the early stage of training, we define $c_i^t$ as the prototype of class $i$ at time step $t$. The initial prototypes $c_i^0$ for each class $i$ are randomly initialized. Similar to the anchors, they are sampled from a standard normal distribution, and subsequently normalized to have unit L2-norm. This ensures that prototypes start in a general manner without initial bias from specific data instances.
	
	To obtain the appropriate representation for each class, we continuously update the prototypes iteratively by the Exponential Moving Average (EMA). At each training step $t$, after computing the features for a mini-batch, for each class $i$ present in that mini-batch, we first calculate $f_{i, \text{avg}}$ as the average of the feature vectors of all samples belonging to class $i$ within this current mini-batch. Then, the prototype $c_i^t$ is updated as:
	\begin{align}
		c_{i}^{t} = \alpha  (c_{i}^{(t-1)}) + (1-\alpha) (f_{i, \text{avg}}),
	\end{align}
	where $ \alpha$ is a momentum coefficient. The prototypes are typically normalized after each update to maintain consistency with the anchors and features, especially if cosine similarity is used as a distance metric in subsequent stages.

	\textbf{Anchor-Prototype Matching.} A prototype $c_i$ is matched with the anchor $v_{\sigma_i}$ with the following constraints:
	\begin{align}
		\boldsymbol{\sigma}^* = \arg\min_{\boldsymbol{\sigma}\in S} \sum_{i=1}^{C}\left \| v_{\sigma_i}-c_i \right \|,  
	\end{align}

	where $S$ denotes the set of all permutation of $\{	1, 2,..., C\}$, and
$\boldsymbol{\sigma} = [\sigma_1, \sigma_2,...,\sigma_C]^\top$ is a permutation vector that assigns the $\sigma_i$-th anchor $v_{\sigma_i}$ to the  prototype $c_i$ of the $i$-th class.   The objective is to minimize the total distance between each class prototype $c_i$ and its matched anchor $v_{\sigma_i}$. This optimization problem seeks to determine the optimal $\boldsymbol{\sigma}^*$. Such a problem can be modeled as an assignment problem and can be efficiently solved by the Hungarian algorithm \cite{jonker1986improving}.

	\textbf{Proximity-Guided Sampling.} The memory bank stores representative features from labeled data and is used for contrastive learning \cite{alonso2021semi,xiao2022semi,wang2022semi}. Due to space limitations, the proposed memory bank only stores a subset of representative features obtained from labeled data, which comprises a first-in-first-out (FIFO) queue for each class. Based on this, we propose a proximity-guided sampling strategy to obtain a more compact memory bank. We extract features only from Gen-Teacher by a projection head. To select the subset of features to be included in the memory bank, we predefine a threshold $ \phi $ to choose the features with higher confidence, and then those features closer to their corresponding anchors are preferred. We update the memory bank only with the $top-K$ closest features and pop out the outdated features at the top.
	
	\textbf{Dual Contrastive Loss.} Traditionally, contrastive learning methods push features toward other features within the positive class and pull them away from others in negative classes. In our method, since features in the memory bank are selected to be close to their corresponding class anchor, we only need to keep features far away from the class anchors that they do not belong to, without needing exhaustive comparisons with all other features. We divide contrastive learning into two parts: anchor contrastive loss ($\mathcal{L}_{ac}$) and similarity loss ($\mathcal{L}_{sim}$). 
	
	Specifically, $\mathcal{L}_{ac}$ brings the feature closer to its corresponding anchor while pushing it farther away from other anchors, forming uniform feature distributions. For each feature $f$, the corresponding class is $c$, which can be computed as follows:
	\begin{align}
		\mathcal{L}_{ac}=-log\frac{exp(f\cdot v_{\sigma _c}/\tau )}{exp(f\cdot v_{\sigma _c}/\tau )+\sum_{c^-\in \sigma}exp(f\cdot v_{\sigma _{c^-}}/\tau) },
\end{align}
where $v_{\sigma _c}$ denotes the corresponding anchor of class $c$, and $v_{\sigma _{c^-}}$ are other class anchors. 

Additionally, $\mathcal{L}_{sim}$ hopes that the features of the same class can be more compact.
\begin{align}
\mathcal{L}_{sim}=
\frac{1}{\left | \mathcal{Q}_c \right | } \sum_{i^+\in \mathcal{Q}_c}(1-\frac{\left \langle f,i^+ \right \rangle }{\left \| f \right \|_2\cdot \left \| i^+ \right \|_2  } ),
\end{align}
where $f$ represents the feature of the predicting pixel belonging to class $c$, $i^+$ denote the representative features of class $c$ in the memory bank $\mathcal{Q}_c$. 

\textbf{Complexity Analysis.} We conduct time and space complexity analysis of dual contrastive learning.
	The dual contrastive objective consists of the anchor contrastive loss $\mathcal{L}_{ac}$ and the similarity loss $\mathcal{L}_{sim}$.
	For $\mathcal{L}_{ac}$, each feature vector $f$ interacts with one positive anchor and $C-1$ negative anchors. With feature dimension $D$, this yields a per-feature cost of $O(CD)$, and processing $N_f$ features per batch results in a total cost of $O(N_f C D)$. The memory required to store the class anchors is $O(CD)$.
	For $\mathcal{L}_{sim}$, each feature compares with $|\mathcal{Q}_c|$ stored samples of its class in the memory bank. Each cosine similarity computation costs $O(D)$, giving a per-feature cost of $O(|\mathcal{Q}_c| D)$ and a per-batch cost of $O(N_f |\mathcal{Q}_c| D)$. The memory footprint of the class-specific memory bank is $O(|\mathcal{Q}_c| D)$.
	


\subsection{Training Process}
In summary, the overall loss for each mini-batch is calculated as follows:
\begin{align}
\mathcal{L}_{total}=\mathcal{L}_{s}+ \lambda_{u}(	\mathcal{L}_{u}^{Gen} + \mathcal{L}_{u}^{Pro})+\lambda_{ctr} (\mathcal{L}_{ac}+\mathcal{L}_{sim}), 
\end{align}
where $\mathcal{L}_{s}$ represents supervised loss, $\mathcal{L}^{Gen}_u$ is general training loss, $\mathcal{L}^{Pro}_u$ is professional training loss, $\mathcal{L}{ac}$ stands for anchor contrastive loss, and $\mathcal{L}{sim}$ refers to similarity loss. The parameters $\lambda_{u}$ and $\lambda_{ctr}$ are weighting coefficients in the overall loss function.

\section{Experimental Results}
In this section, we evaluate the effectiveness of our proposed method on two widely-used semantic segmentation datasets for autonomous driving: the Cityscapes dataset \cite{cordts2016cityscapes} and the CamVid dataset \cite{brostow2008segmentation}. In addition, to compare our approach with state-of-the-art methods, we also conduct performance evaluations on the PASCAL VOC 2012 dataset. The experiments are implemented using PyTorch on a server with an NVIDIA A40 GPU.

\subsection{Experimental Setting}
\textbf{Datasets.}
There are two datasets for semantic segmentation in urban driving scenarios: The Cityscapes dataset \cite{cordts2016cityscapes}, designed for semantic analysis of urban street scenes, contains 2975 finely annotated training images and 500 validation images across 19 classes, while the CamVid dataset \cite{brostow2008segmentation}, captured from a driving automobile's perspective, includes 367 training and 101 validation images across 11 classes; however, both datasets exhibit a significant long-tail distribution, where dominant classes like roads and sky occupy the majority of pixels, whereas underrepresented classes such as pedestrians and poles comprise much fewer pixels, severely hindering model learning performance. In addition, to compare our approach with state-of-the-art methods, we also verified the effectiveness of our approach on PASCAL VOC 2012 dataset which \cite{everingham2010pascal} serves as a standard benchmark for semantic segmentation with 20 object classes and 1 background class. And it includes 1464 training images and 1449 validation images. Additionally, we use the SBD dataset \cite{hariharan2011semantic} as an augmented set, which provides 9118 extra training images. Due to the coarse annotations in the SBD dataset, previous methods employ two primary partitioning strategies. The first strategy uses only the standard 1464 images as the entire labeled set, while the second strategy considers all 10582 images as potential labeled data. To ensure a fair comparison of our methods, we evaluate them using both a blender set (10582 potential labeled images) \cite{wang2022semi} and a classic set (1464 labeled images) \cite{chen2021semi}.

\textbf{Evaluation.}
To ensure a fair comparison with prior work, we use DeepLabv3+ \cite{chen2018encoder} pretrained on ImageNet \cite{deng2009imagenet} as our segmentation model. The projection head for contrastive learning comprises a Conv-ReLU-Dropout-Conv block. The segmentation head and projection head map the ASPP output to $C$ classes and a 256-dimensional feature space, respectively. Consistent with previous studies, we evaluate segmentation performance using the mean intersection-over-union (mIoU) metric across all datasets. We utilize Gen-Student to evaluate the performance of our novel framework.

\textbf{Implementation Details.}
Prior to the main training phase, we pre-optimize for 50 epochs using an Adam optimizer and a $10^{-3}$ learning rate to generate evenly distributed anchors.
The two student models share the same architecture, utilizing DeepLabV3+, but have different initializations. Both models are trained using the stochastic gradient descent (SGD) optimizer. When training on the Cityscapes and CamVid dataset, the learning rate is initialized to ${10^{-2}}$, whereas, for the PASCAL VOC 2012 dataset, it is $5 \times 10^{-3}$. The momentum for the optimizer is maintained at $0.9$. For adjusting the learning rate, we employ a polynomial decay strategy defined by $1-(\frac{\text{iter}}{\text{max\_iter}})^{\text{power}}$, with the power parameter set to $0.9$.
During training, images are randomly cropped to $800 \times 800$ pixels with a batch size of 4 for the Cityscapes dataset. For the PASCAL VOC 2012 dataset, a crop size of $512 \times 512$ pixels with a batch size of 8 is used. In the case of the CamVid dataset, the configuration remains similar to that of Cityscapes, except that the images are cropped to $360 \times 480$ pixels and the batch size is set to 16. The weights and hyperparameters for the loss functions are configured as follows: $\lambda_{s} = 1$, $\lambda_{u} = 1$, $\lambda_{ctr} = 0.1$, $\tau = 0.5$, $ \phi = 0.95$, and $N = 256$. To implement strong data augmentation, three rectangular regions with random ratios ranging from 0.25 to 0.5 are randomly positioned within the input image and augmented using the CutMix \cite{yun2019cutmix} strategy.

\subsection{Results}
\textbf{Results on the Cityscapes dataset.} We evaluated the effectiveness of STPG using the Cityscapes dataset. To ensure fairness, DeepLabV3+ with ResNet50 is employed across all comparison methods. In all experiments, we arbitrarily choose 1/8, 1/4, and 1/2 of the Cityscapes training set as labeled data, corresponding to 372, 744, and 1488 images respectively, with the remaining portion as unlabeled data. Table \ref{city} presents the comparative results, where STPG demonstrates a substantial improvement over existing methods. Specifically, STPG enhances the mIoU by 8.40\%, 5.85\% and 4.12\% in the 1/8, 1/4, and 1/2 columns, respectively, compared to baseline methods using only labeled data.

\begin{table}[!htbp]
	\caption{Comparisons of STPG with the state-of-the-art methods on the Cityscapes validation set using various partition protocols. We arbitrarily choose subsets from the training set of Cityscapes to use as labeled data: 1/8, 1/4 and 1/2, corresponding to 372, 744, and 1488 images respectively, while the remaining training data serves as unlabeled. Each method employs DeepLabV3+ with ResNet50.  $*$ indicates a reproduction under the same settings as ours.}
	\centering
	\label{city}
	\resizebox{9cm}{!}{
		\begin{tabular}{l|c|ccccc}  
			\toprule   
			\textbf{Methods} & \textbf{Year}  & 1/8 (372) & 1/4 (744) & 1/2 (1488) \\  
			\midrule
			Sup. baseline & -  & 66.74 & 71.16 & 74.36 \\
			MT \cite{tarvainen2017mean} & 2017  & 72.0 & 74.5&-\\
			CCT \cite{ouali2020semi} & 2020  & 72.5 & 75.7&-\\
			GCT \cite{ke2020guided} & 2020  & 71.3 & 75.3&-\\
			CAC \cite{lai2021semi} & 2021  & 69.7 & 72.7&-\\
			PC$^2$Seg \cite{zhong2021pixel} & 2021  & 72.1 & 73.8&- \\
			ELN \cite{kwon2022semi} & 2022  & 70.3 & 73.5 &75.3 \\
			ST++ \cite{yang2022st++} & 2022   & 72.7 & 73.8 &- \\
			PGCL \cite{kong2023pruning} & 2023  & 71.2 & 73.9 & 76.8   \\
			MMFA \cite{yin2023semi} & 2023  & 73.6 & 76.9&- \\
			CPCL \cite{fan2023conservative} & 2023  & 74.6 & \underline{76.9} &-\\
			UniMatch$^*$ \cite{yang2023revisiting} & 2023 & \textbf{75.7} & 76.8 & \underline{78.1} \\
			MGCT \cite{hu2024multi} & 2024  & 73.4 & 74.0&- \\
			CorrMatch$^*$ \cite{sun2024corrmatch} & 2024 & 74.1 & 76.8&77.1 \\
			UCCL$^*$ \cite{yin2025uncertainty} & 2025 & 71.2 & 73.1&74.3 \\
			ScaleMatch$^*$ \cite{lv2025scalematch} & 2025 & 75.1 & 76.5&76.9 \\
			NRCR$^*$ \cite{zhang2025noise} & 2025 & 73.5 & 73.9&75.8 \\
			MCCL$^*$ \cite{yin2025semi} & 2025 & 71.0 & 70.6&72.7 \\
			\midrule
			\textbf{STPG(Ours)} & 2025& \underline{75.14} & \textbf{77.01} & \textbf{78.48} \\
			\bottomrule  
		\end{tabular}
	}
\end{table}

\begin{figure}[!h]
	\begin{center}
		\captionsetup[subfloat]{labelsep=none,format=plain,labelformat=empty}
		\subfloat{\includegraphics[width=1\linewidth]{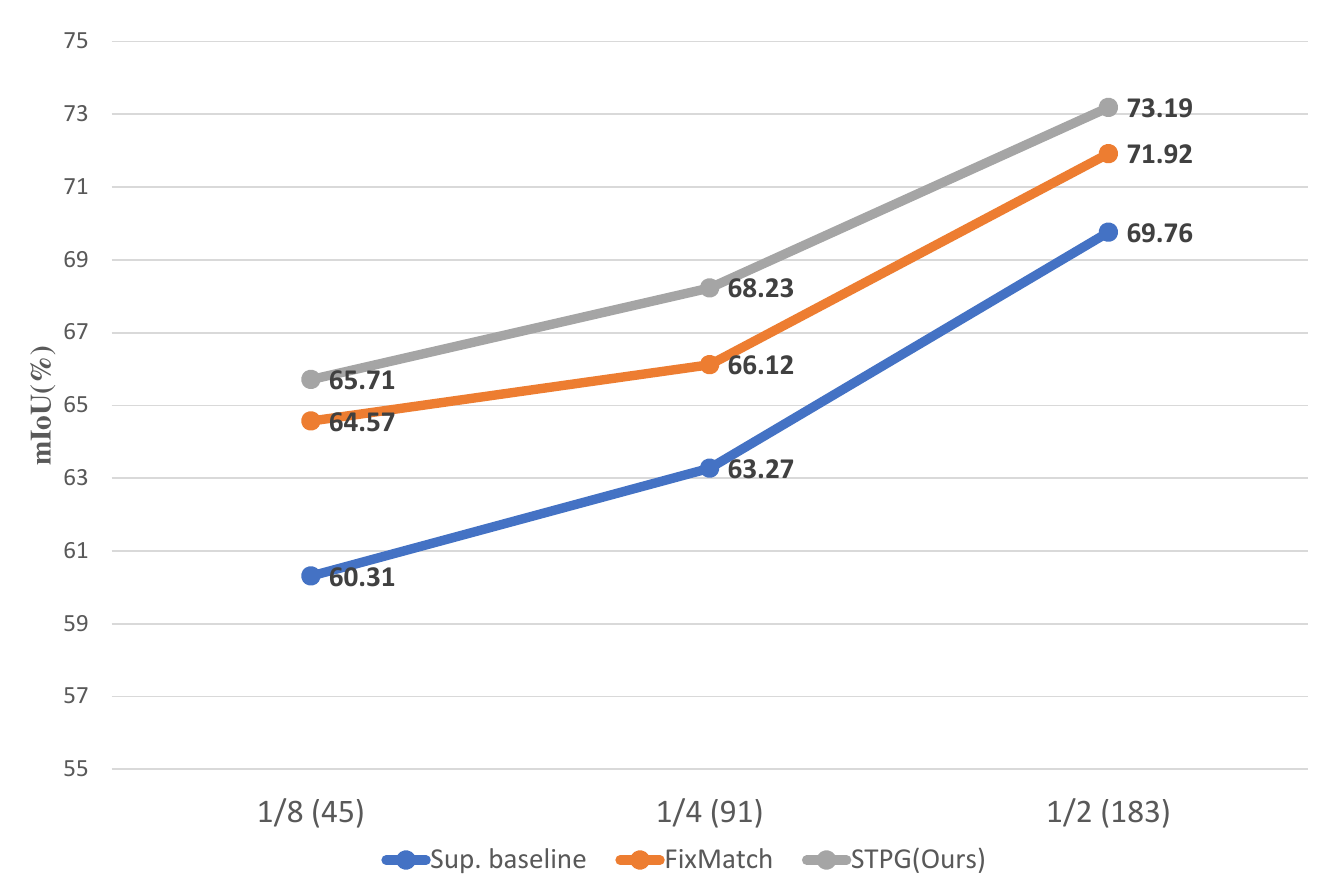}} \\	
	\end{center}
	\vspace{-10pt}
	
	\caption{\textbf{Performance comparison on the CamVid dataset.} Comparisons of STPG with baseline and FixMatch on the CamVid validation set using various partition protocols. We arbitrarily choose subsets from the CamVid training set to use as labeled data: 1/8, 1/4, and 1/2, corresponding to 45, 91, and 183 images respectively, while the rest of the training set serves as unlabeled data. Each method employs DeepLabV3+ with ResNet50.
	}
	\label{camvid}
	\vspace{-10pt}
	
\end{figure}

\textbf{Results on the CamVid dataset.} We evaluated the effectiveness of STPG using the CamVid dataset. To ensure fairness, DeepLabV3+ with ResNet50 is employed. In all experiments, we arbitrarily choose 1/8, 1/4, and 1/2 of the training set as labeled data, corresponding to 45, 91, and 183 images respectively, with the remaining portion as unlabeled data. Figure \ref{camvid} presents the comparative results, where STPG demonstrates a substantial improvement over existing methods. Specifically, STPG enhances the mIoU by 5.40\%, 4.96\%, and 3.43\% in the 1/8, 1/4, and 1/2 columns, respectively, compared to baseline methods using only labeled data.

\textbf{Results on the \emph{blender} PASCAL VOC 2012 Dataset. } We further assessed the performance of STPG on the blender PASCAL VOC 2012 dataset, ensuring all comparison methods utilized DeepLabV3+ with ResNet50 for consistency. In all experiments, we arbitrarily choose 1/16, 1/8, 1/4, and 1/2 of the PASCAL VOC 2012 training set as labeled data, corresponding to 662, 1323, 2646, and 5291 images respectively, with the remaining images treated as unlabeled data. Table \ref{voc} presents the comparative results. STPG consistently achieves state-of-the-art results across all partition protocols, surpassing existing best results by 0.08\%, 0.19\%, 1.17\%, and 1.19\% for the respective data splits.

\textbf{Results on the \emph{classic} PASCAL VOC 2012 Dataset. } We evaluate STPG on the classic PASCAL VOC 2012 dataset by selecting labeled images from the original fine-grained annotated training set (1464 images total) and treating all images from the SBD as unlabeled data. For fairness, all comparative methods employed DeepLabV3+ with ResNet101. In all experiments, we arbitrarily choose 1/8, 1/4, 1/2, and the full set of the PASCAL VOC 2012 training data as labeled, corresponding to 183, 366, 732, and 1464 images respectively, with the remainder as unlabeled. Table \ref{voc101} shows the comparison results. STPG consistently outperforms the supervised-only baseline, with improvements of +17.97\%, +15.38\%, +11.72\%, and +8.33\% for the 1/16, 1/8, 1/4, and 1/2 partition protocols, respectively.

\begin{table}[!h]
	\caption{Comparisons of STPG with the state-of-the-art methods on the \textbf{\emph{blender}} PASCAL VOC 2012 validation set using various partition protocols. We arbitrarily choose subsets from the PASCAL VOC 2012 training set to use as labeled data: 1/16, 1/8, 1/4, and 1/2, corresponding to 662, 1323, 2646, and 5291 images respectively, while the rest of the training set serves as unlabeled data. Each method employs DeepLabV3+ with ResNet50. $*$ indicates a reproduction under the same settings as ours.}
	\centering
	\label{voc}
	\resizebox{9cm}{!}{
		\begin{tabular}{l|c|ccccc}
			\toprule   
			\textbf{Methods} & \textbf{Year} & 1/16 (662)  & 1/8 (1323) & 1/4 (2646) &1/2 (5291) \\  
			\midrule
			Sup. baseline & - & 61.72 & 66.83 & 71.52 & 73.96 \\
			CutMix-Seg \cite{french2019semi} & 2019 & 68.9 & 70.7 &  72.5 & 74.5\\
			CPS \cite{chen2021semi} & 2021 & 72.0 & 73.7 & 74.9 & 76.2  \\
			ADS-SemiSeg \cite{cao2022adversarial} & 2022 & - & 70.8 &  72.8 & -\\
			ELN \cite{kwon2022semi} & 2022 & - & 73.2 & 75.6 & -\\
			ST++ \cite{yang2022st++} & 2022 & 72.6 & 74.4 & 75.4  & -\\		
			U$^2$PL \cite{wang2022semi} & 2022 & 72.0 & 75.1 & 76.2 & - \\
			CPCL \cite{fan2023conservative} & 2023 & 71.7 & 73.7 & 74.6 & 75.3 \\
			PGCL \cite{kong2023pruning} & 2023 & - & 75.2 & 76.0 & - \\
			MMFA \cite{yin2023semi} & 2023 & 68.0& 72.5 & 75.7 & - \\
			UniMatch$^*$ \cite{yang2023revisiting} & 2023 & 73.4 & 75.1 &\underline{76.1} & \underline{76.9} \\
			VC$^3$ \cite{hou2024view} & 2024 & \underline{73.8}&75.3&75.8&-\\
			MGCT \cite{hu2024multi} & 2024 & 73.4&\textbf{75.6}&75.9&-\\
			\midrule
			\textbf{STPG(Ours)}  & 2025 & \textbf{73.88} & \underline{75.49} & \textbf{76.97} & \textbf{77.39} \\
			\bottomrule  
		\end{tabular}
	}
\end{table}

\begin{figure}[t]
\begin{center}
	\captionsetup[subfloat]{labelsep=none,format=plain,labelformat=empty}
	\subfloat{\includegraphics[width=1\linewidth]{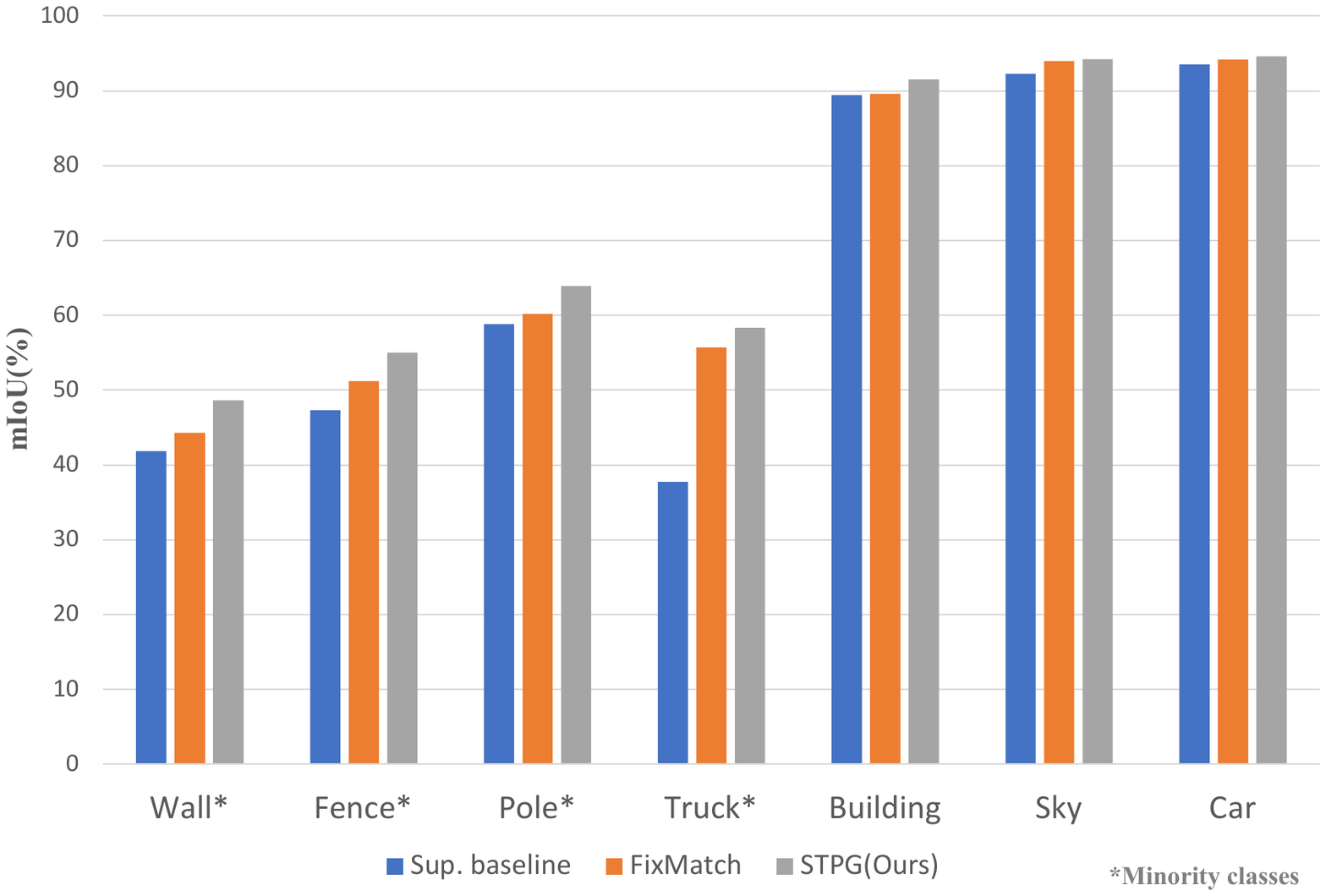}} \\	
\end{center}
\vspace{-15pt}

\caption{\textbf{Enhanced performance for different classes.} The mIoU(\%) for the tail classes (such as Wall, Fence, Pole, etc.) and majority classes (such as Building, Sky, and Car), was evaluated using the baseline, FixMatch, and STPG. The results are derived from the Cityscapes dataset using 1/8 of the labeled data, employing DeepLabV3+ with ResNet50.
}
\label{cmp}
\vspace{-12pt}

\end{figure}

\begin{table}[!htbp]
	\caption{Comparisons of our STPG with the state-of-the-art methods on the \textbf{\emph{classic}} PASCAL VOC 2012 validation set using different partition protocols. We arbitrarily choose subsets from the fine-grained annotated training set of PASCAL VOC 2012 to use as labeled data: 1/8, 1/4, 1/2, and the full set, corresponding to 183, 366, 732, and 1464 images respectively. The remaining training set, including all SBD images, serves as unlabeled data. Each method employs DeepLabV3+ with ResNet101. $*$ indicates a reproduction under the same settings as ours.}
	\centering
	\label{voc101}
	\resizebox{9cm}{!}{
		\begin{tabular}{l|c|ccccc}
			\toprule   
			\textbf{Methods} & \textbf{Year} & 1/8 (183) & 1/4 (366) & 1/2 (732) & Full (1464)\\  
			\midrule   
			Sup. baseline & - & 54.32 & 62.31 & 67.36 & 72.12 \\
			CutMix-Seg \cite{french2019semi} & 2019 & 63.5 & 69.5 & 73.7 & 76.5 \\
			PseudoSeg \cite{zou2020pseudoseg} & 2020 & 65.5 & 69.1 & 72.4 & 73.2 \\
			PC$^2$Seg \cite{zhong2021pixel} & 2021 & 66.3 & 69.8 & 73.1 & 74.2 \\
			CPS \cite{chen2021semi} & 2021 & 67.4 & 71.7 & 75.9 & - \\
			CTT \cite{xiao2022semi} & 2022 & 71.1 & 72.4 & 76.1 & - \\
			ST++ \cite{yang2022st++} & 2022 & 71.0 & 74.6 & 77.3 & 79.1 \\
			U$^2$PL \cite{wang2022semi} & 2022 & 69.2 & 73.7 & 76.2 & 79.5 \\
			PS-MT \cite{liu2022perturbed} & 2022 & 69.6 & 76.6 & 78.4 & 80.0 \\
			FPL \cite{qiao2023fuzzy} & 2023 & 71.7 & 75.7 & 79.0 & - \\
			PRCL \cite{xie2023boosting} & 2023 & \textbf{74.4} & 76.7 & - & 78.2 \\
			SemiCVT \cite{huang2023semicvt} & 2023 & 71.3 & 75.0 & 78.5 & 80.3 \\
			UPC \cite{fang2023locating} & 2023 & 73.5 & 76.1 & 78.0 & 80.2 \\
			CSS \cite{wang2023space} & 2023 & 71.9 & 74.9 & 77.6 & - \\
			UniMatch$^*$ \cite{yang2023revisiting} & 2023 & \underline{74.1} & \underline{77.4} & \underline{79.0} & 80.1 \\
			MGCT \cite{hu2024multi} & 2024 & 73.8 & 77.1 & 78.2 & \underline{80.3} \\
			\midrule
			\textbf{STPG(Ours)} & 2025 & 72.29 & \textbf{77.69} & \textbf{79.08} & \textbf{80.45} \\
			\bottomrule  
		\end{tabular}
	}
\end{table}
\textbf{Enhanced Performance for Tail Classes.}   Due to the significant long-tail problem in the Cityscapes dataset, we conducted an additional evaluation on some of the classes, especially the minority classes (such as Wall, Fence, Pole, etc.), and compared STPG with the baseline and FixMatch. As shown in Fig. \ref{cmp}, STPG can significantly improve the performance of the model on minority classes. Furthermore, by enhancing the learning of minority classes and introducing anchor-based contrastive learning, our framework enables the model to achieve better separation between minority and majority classes in the feature space. This improved feature separation not only benefits minority classes but also helps majority class samples to cluster more tightly in their respective regions, leading to marginal performance gains for majority classes as well. This observation highlights the strength of our framework: it effectively enhances the recognition performance of minority classes without compromising the accuracy of majority classes. 

\begin{figure}[t]
\begin{center}
	\captionsetup[subfloat]{labelsep=none,format=plain,labelformat=empty}
	\subfloat{\includegraphics[width=1\linewidth]{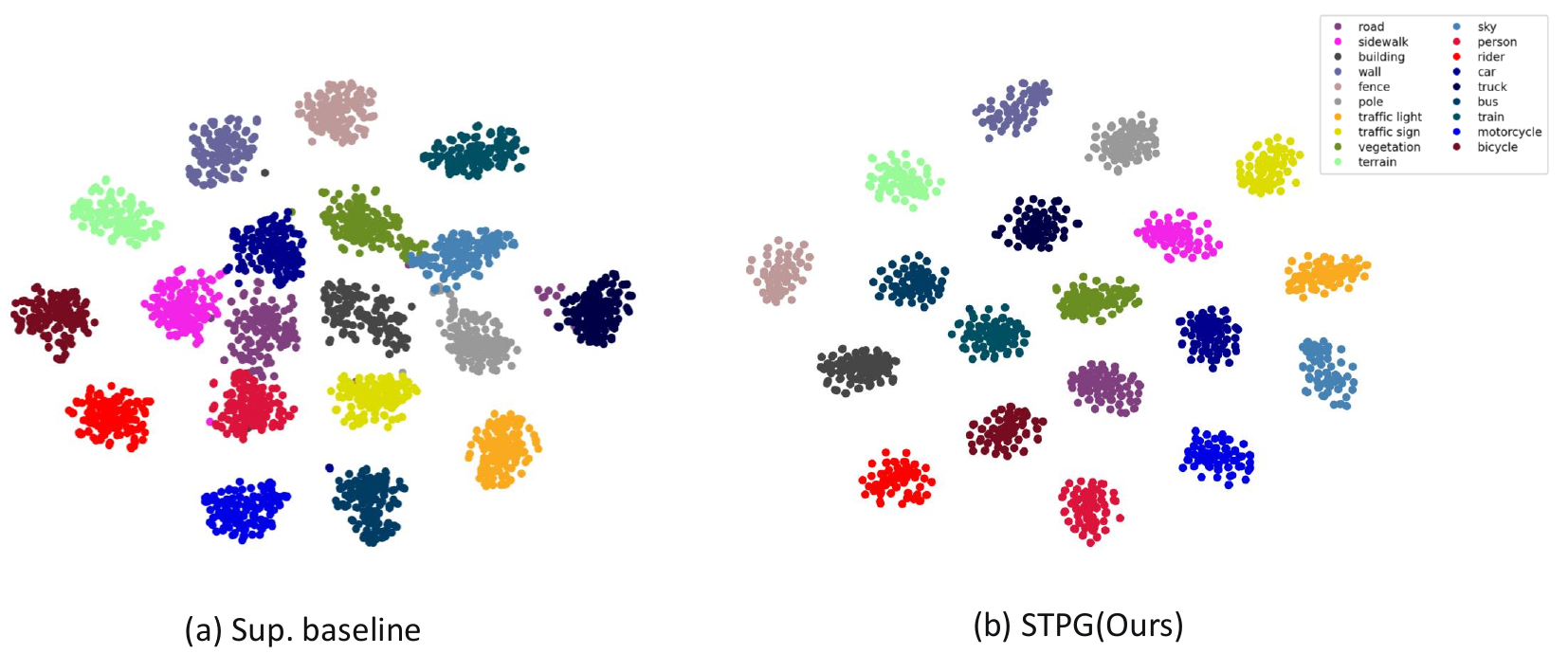}} \\	
\end{center}
\vspace{-15pt}

\caption{\textbf{Comparison of T-SNE visualization.} We use T-SNE to project features derived from the input data into a 2D space. For visualization, we sample 256 points per class. The resulting plot illustrates that STPG achieves superior clustering performance.
}
\label{tsne}
\vspace{-8pt}

\end{figure}

\textbf{Comparison of T-SNE Visualization.}
We compare the data feature distributions produced by the
baseline and STPG after T-SNE visualization. In Fig. \ref{tsne}, the T-SNE visualization of the model output features reveals that, compared to the baseline where feature distributions are scattered and class boundaries are blurred, our method demonstrates superior performance in feature alignment and clustering, with data points forming distinct class clusters and achieving better separation between classes; this observation indicates that our method can more effectively learn discriminative feature representations, leading to more accurate predictions.

\subsection{Analysis}
In this section, we assess the performance improvements of proposed modules by conducting a series of experiments. These experiments use the Cityscapes and blender PASCAL VOC 2012 datasets, each with 1/8 of the data labeled. We utilize DeepLabV3+ with ResNet50 for all experiments.

%

\begin{table}[h]
\caption{Improvement of each proposed module: $\mathcal{L}_{s}$ represents the supervised loss, while $\mathcal{L}_{u}$ denotes the unsupervised loss. DM signifies a dual mean teacher architecture, PS denotes the pixel selection strategy, and ACL indicates anchor contrastive learning. The results are obtained using 1/8 labeled data from both the Cityscapes and PASCAL VOC 2012 datasets, employing DeepLabV3+ with ResNet50.}
\centering
\label{module}
\scalebox{0.9}{
	\begin{tabular}{ccccc|cc}
		\toprule 
		\multicolumn{5}{c|}{}                                                          & \multicolumn{2}{c}{mIoU(\%)}                                           \\ \midrule  
		$\mathcal{L}_s$ & $\mathcal{L}_u$ & DM          & PS           & ACL         & Cityscapes & \begin{tabular}[c]{@{}c@{}}PASCAL VOC\\ 2012\end{tabular} \\ \midrule
		$\checkmark$    &                 &              &              &              & 66.74      & 66.83                                                         \\
		$\checkmark$    & $\checkmark$    &              &              &              & 72.10      & 72.63                                                         \\
		$\checkmark$    & $\checkmark$    & $\checkmark$ &              &              & 73.37      & 73.92                                                        \\
		$\checkmark$    & $\checkmark$    & $\checkmark$ & $\checkmark$ &              & 73.96      & 74.87                                                        \\
		$\checkmark$    & $\checkmark$    & $\checkmark$ &              & $\checkmark$  & 74.19      & 74.26                                                        \\
		$\checkmark$    & $\checkmark$    & $\checkmark$ & $\checkmark$ & $\checkmark$ & \textbf{75.14 }     & \textbf{75.49}                                                         \\ \bottomrule 
\end{tabular}}
\end{table}

\textbf{Ablation Studies.} Table \ref{module} presents the contributions of each module within our framework. Each module contributes significantly to enhancing semi-supervised semantic segmentation performance. Initially, the baseline $\mathcal{L}_{s}$, which is trained solely on labeled data, achieves a certain mIoU score. Then, we adopt the traditional mean-teacher architecture to introduce unlabeled data $\mathcal{L}_{u}$ and improve the performance of the model. However, the introduction of the dual mean teacher architecture (DM) individually boosts the mIoU to 73.37\% for Cityscapes and 73.92\% for PASCAL VOC 2012, indicating improvements of 6.63\% and 7.09\%, respectively. Furthermore, incorporating the pixel selection strategy (PS) and anchor contrastive learning (ACL) modules leads to additional performance gains. The PS and ACL modules surpass the baseline by 7.22\% and 7.45\%, respectively, for Cityscapes and by 8.04\% and 7.43\%, respectively, for PASCAL VOC 2012. Upon integrating these modules, the overall performance escalates, reaching 75.14\% for Cityscapes and 75.49\% for PASCAL VOC 2012. This comprehensive analysis underscores the effectiveness of the dual mean teacher architecture, pixel selection strategy, and anchor contrastive learning in enhancing semi-supervised semantic segmentation.

{\textbf{Effect of Pixel Selection Strategy.} As shown in Table \ref{pseudo}, our ablation study highlights the critical role of pseudo-label selection. A baseline using only high-quality, consistently predicted pseudo-labels establishes a solid foundation but achieves a modest mIoU of 71.43\%, as detailed in Row (a). This suggests that relying exclusively on easy examples limits model generalization. The key to our improved performance lies in strategically leveraging uncertainty. By incorporating high-mismatch pseudo-labels, which are identified at the class level, alongside the high-quality ones, performance significantly increases to a peak of 75.14\% mIoU in Row (b). Adding high-mismatch pseudo-labels consistently boosts performance, improving mIoU by +3.71\% in Row (b) and +0.95\% in Row (d). This result validates our central hypothesis: uncertain, boundary-defining samples are more valuable for model improvement than deterministic ones alone. We posit that these high-mismatch pseudo labels, often generated for hard-to-classify minority classes, are rich in valuable boundary information. This is because these classes are inherently prone to being misclassified as more dominant, over-confident classes, thus yielding higher mismatch scores. Conversely, other configurations prove less effective. While adding low-mismatch labels shows a slight improvement over the baseline in Row (c), combining both low- and high-mismatch labels in Row (d) leads to a performance drop relative to our best result. This demonstrates that it is specifically the high-mismatch samples that drive the improvement.
}

\begin{table}[h]

\setlength\tabcolsep{15pt}  
\caption{Effect of pixel selection strategy. $\hat{y}^{Hq}_u$ indicates high-quality pseudo-labels. $\hat{y}^{Lmis}_u$ indicates low-mismatch pseudo-labels. $\hat{y}^{Hmis}_u$ indicates high-mismatch pseudo-labels. The results are derived from the Cityscapes dataset using 1/8 of the labeled data, employing DeepLabV3+ with ResNet50.}
\centering
\label{pseudo}
\scalebox{0.75}{
	\begin{tabular}{c|ccc|cc} 
		\toprule   
		& $\hat{y}^{Hq}_u$ & $\hat{y}^{Lmis}_u$ & $\hat{y}^{Hmis}_u$ & $mIoU(\%)$ & $\Delta_{mIoU}(\%)$ \\ 
		\midrule
		(a) & \checkmark & &  &  71.43 & -- \\ 
		(b) & \checkmark & & \checkmark &  \textbf{75.14} & +3.71 \\ 
		\midrule
		(c) & \checkmark & \checkmark &  &  72.61 & -- \\
		(d) & \checkmark & \checkmark & \checkmark & 73.56 & +0.95 \\
		\bottomrule  
	\end{tabular}
}

\end{table}

\begin{figure}[t]
	\begin{center}
			\captionsetup[subfloat]{labelsep=none,format=plain,labelformat=empty}
			\subfloat{\includegraphics[width=1\linewidth]{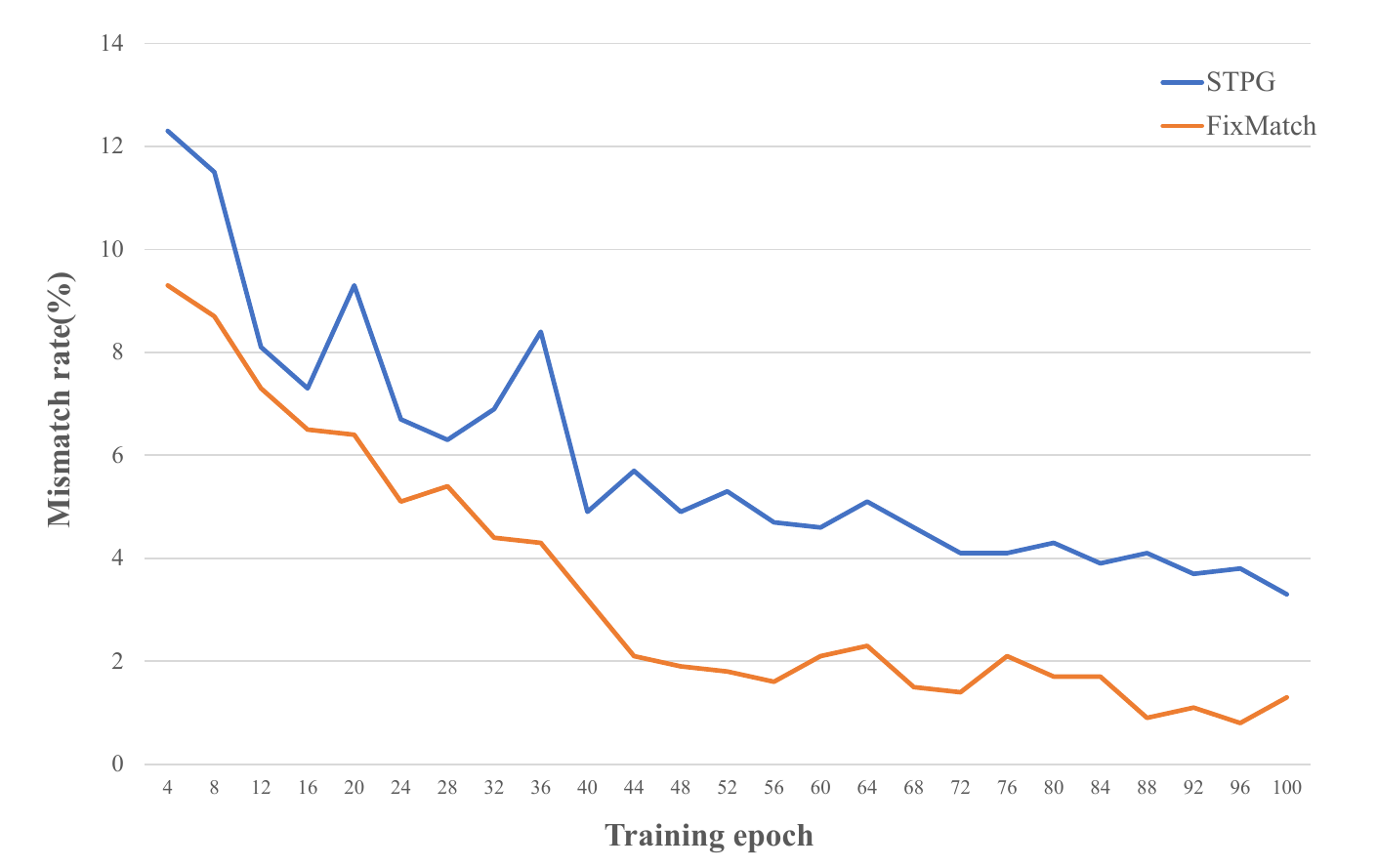}} \\	
		\end{center}
	\vspace{-15pt}
	
	\caption{\textbf{Mismatch Rate comparison between STPG and FixMatch.} 
		The mismatch rate between the student and teacher models was measured every 4 epochs during training for both methods. The plot indicates that STPG consistently exhibits a higher mismatch rate, suggesting a weaker coupling between its student and teacher models compared to FixMatch.
		}
	\label{disagree}
	\vspace{-22pt}
	
\end{figure}

{\textbf{Model Decoupling Analysis.} 
	To rigorously characterize the divergence between the student and teacher models, we define the mismatch rate $\mathrm{MR}$ as the proportion of evaluation samples $x_i$ whose classification predictions differ between the two models:		
	\begin{align}
		\mathrm{MR} = 
		\frac{1}{N} \sum_{i=1}^{N} \mathbb{I}(\text{argmax} f_\theta(x_i) \neq \text{argmax} f_\xi(x_i)),
	\end{align}
	where $N$ is the total number of evaluate samples. 
	
	As shown in Fig. \ref{disagree}, during training, we periodically measured the mismatch rate between the student and teacher models for both STPG and FixMatch, taking measurements every 4 epochs. The generated plot shows that STPG maintains a consistently higher mismatch rate, revealing a looser student–teacher coupling than that of FixMatch. This higher mismatch rate is consistent with STPG’s design, where cross-teacher supervision and the complementary objectives of the two students reduce parameter coupling and encourage more diverse feature representations.}

\begin{figure}[t]
\begin{center}
	\captionsetup[subfloat]{labelsep=none,format=plain,labelformat=empty}
	\subfloat{\includegraphics[width=1\linewidth]{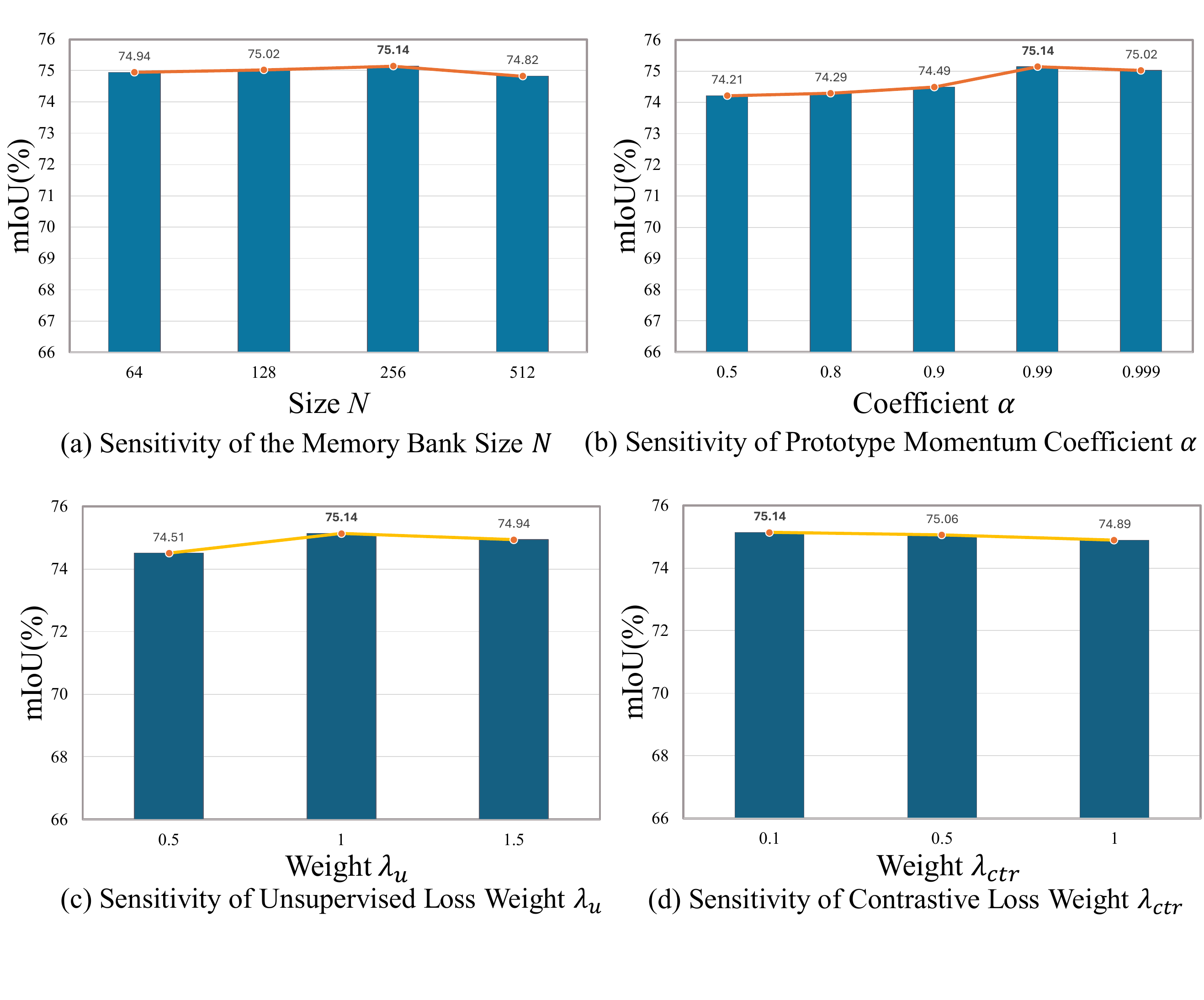}} \\
\end{center}
\vspace{-15pt}

\caption{\textbf{Sensitivity of hyper-parameters.} (a) Sensitivity of the memory bank size $N$, (b)Sensitivity of Prototype Momentum Coefficient $\alpha$. (c) Sensitivity of unsupervised loss weight $\lambda_{u}$, (d) Sensitivity of Contrastive Loss Weight  $\lambda_{ctr}$.
	 The results are derived from the Cityscapes dataset using 1/8 of the labeled data, employing DeepLabV3+ with ResNet50.
}
\label{zhexian}
\vspace{-12pt}

\end{figure}

\textbf{Sensitivity of the Memory Bank Size $N$. }
As illustrated in Fig. \ref{zhexian}(a), the performance of the model is not significantly affected by variations in the size of the memory bank when other settings are kept constant. Minor adjustments to the memory bank size can result in slight changes in model performance. However, using an excessively large memory bank, such as a size of 512, may degrade performance. Since all features in the memory bank are utilized during the training, the computational load is directly proportional to the size $N$. To balance performance and computational efficiency, we have set the default memory bank size to 256.

{
	\textbf{Sensitivity of Prototype Momentum Coefficient $\alpha$.}
Fig.~\ref{zhexian}(b) presents the sensitivity of the model to the prototype momentum coefficient $\alpha$. The results show that the model maintains stable performance across a broad range of $\alpha$ values. When $\alpha$ is between 0.9 and 0.999, the mIoU remains consistently high, varying from 74.49\% to 75.14\%, with the best score obtained at $\alpha=0.99$. Lower values such as $\alpha=0.5$ and $\alpha=0.8$ lead to slight decreases in mIoU, yielding 74.21\% and 74.29\% respectively. These observations indicate that the method is generally robust to the choice of $\alpha$, while a moderately large momentum around 0.99 provides the most reliable performance. }

\textbf{Sensitivity of Loss Weight $\lambda_{u}$ and $\lambda_{ctr}$.} 
We comprehensively explore the impact of both the unsupervised loss weight  $\lambda_{u}$
and the contrastive loss weight $\lambda_{ctr}$
on the final segmentation performance. The results are presented in Fig. \ref{zhexian}(c) and Fig. \ref{zhexian}(d) respectively. For $\lambda_{u}$, when 
$\lambda_{u}$
is set to 0.5, reaching an mIoU of 74.51\%. The optimal performance occurs at 
$\lambda_{u}=1.0$, where the mIoU peaks at 75.14\%. This suggests that an equal emphasis on unsupervised and supervised losses is most beneficial within this experimental setup. However, further increasing $\lambda_{u}$
to 1.5 leads to a slight decrease in performance to 74.94\%, indicating that excessive weighting of unsupervised loss can eventually introduce diminishing returns.
Regarding 
$\lambda_{ctr}$, introducing a small weight of 
$\lambda_{ctr}=0.1$ significantly boosts the performance to 75.14\%, matching the highest mIoU observed. As $\lambda_{ctr}$ is further increased to 0.5, the mIoU remains robust at 75.06\%. Even at $\lambda_{ctr}=1.0$, it holds strong at 74.89\%. This demonstrates that contrastive learning consistently contributes positively to segmentation performance across a range of weights, with $\lambda_{ctr}=0.1$ appearing to be the sweet spot for maximum benefit.
In summary, both unsupervised learning and contrastive learning are crucial components for enhancing segmentation performance. The model shows a clear sensitivity to 
$\lambda_{u}$, with an optimal point empirically found at 1.0. Conversely, the model is relatively robust to the choice of $\lambda_{ctr}$within the tested range, indicating that even a modest incorporation of contrastive loss provides significant and consistent benefits.

\begin{figure}[!htbp]

\begin{center}
	
	\includegraphics[width=1\columnwidth]{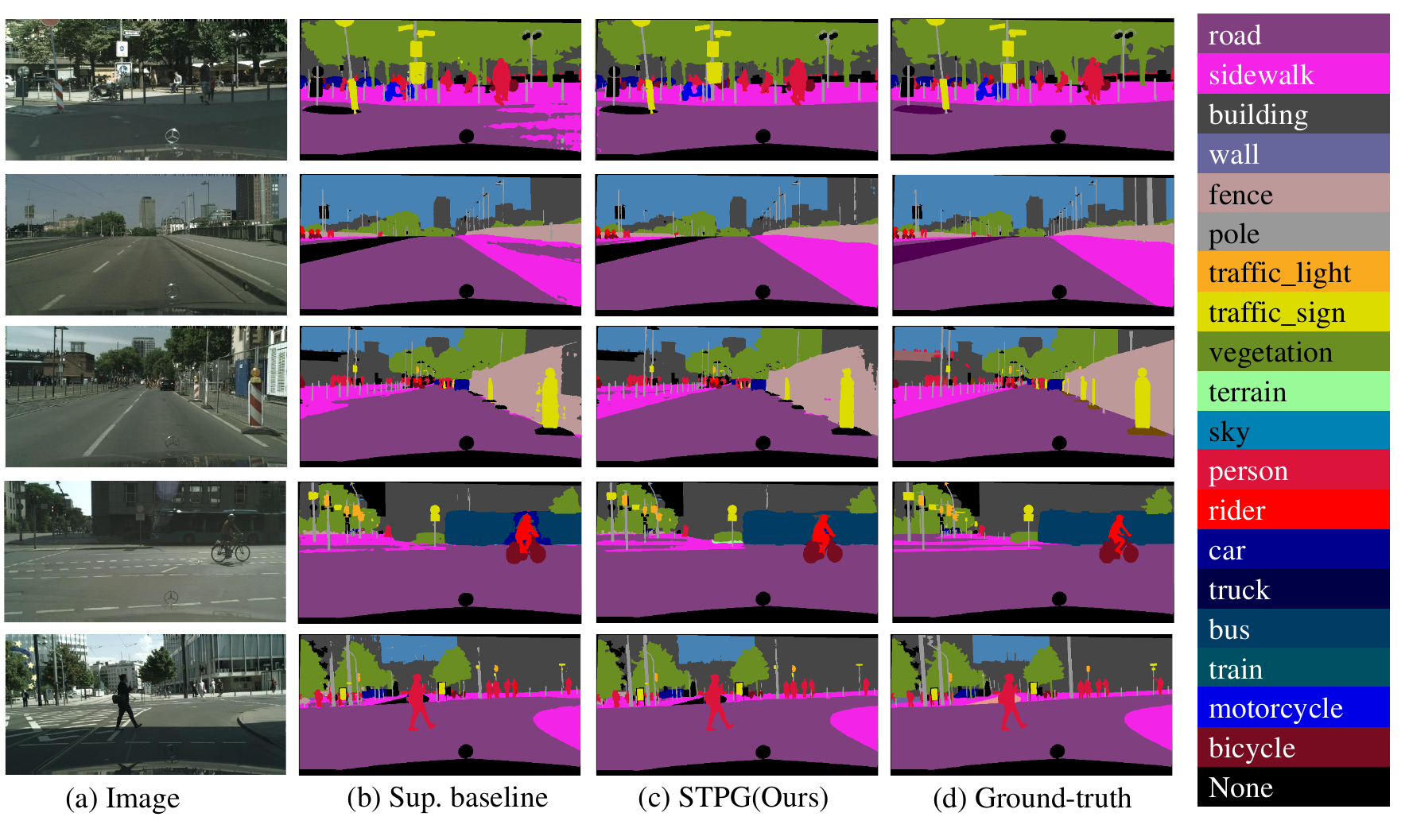}
	
\end{center}

\caption{Qualitative results of STPG framework using 1/8 labeled data from the Cityscapes training set. (a) Images from the validation set. (b) The results of training exclusively with labeled data. (c) The results achieved with STPG. (d) Ground-truths. All experiments are conducted using DeepLabV3+ with ResNet50.}
\label{cityimage}

\end{figure}

\begin{figure}[!htbp]

\begin{center}
	
	\includegraphics[width=0.96\columnwidth]{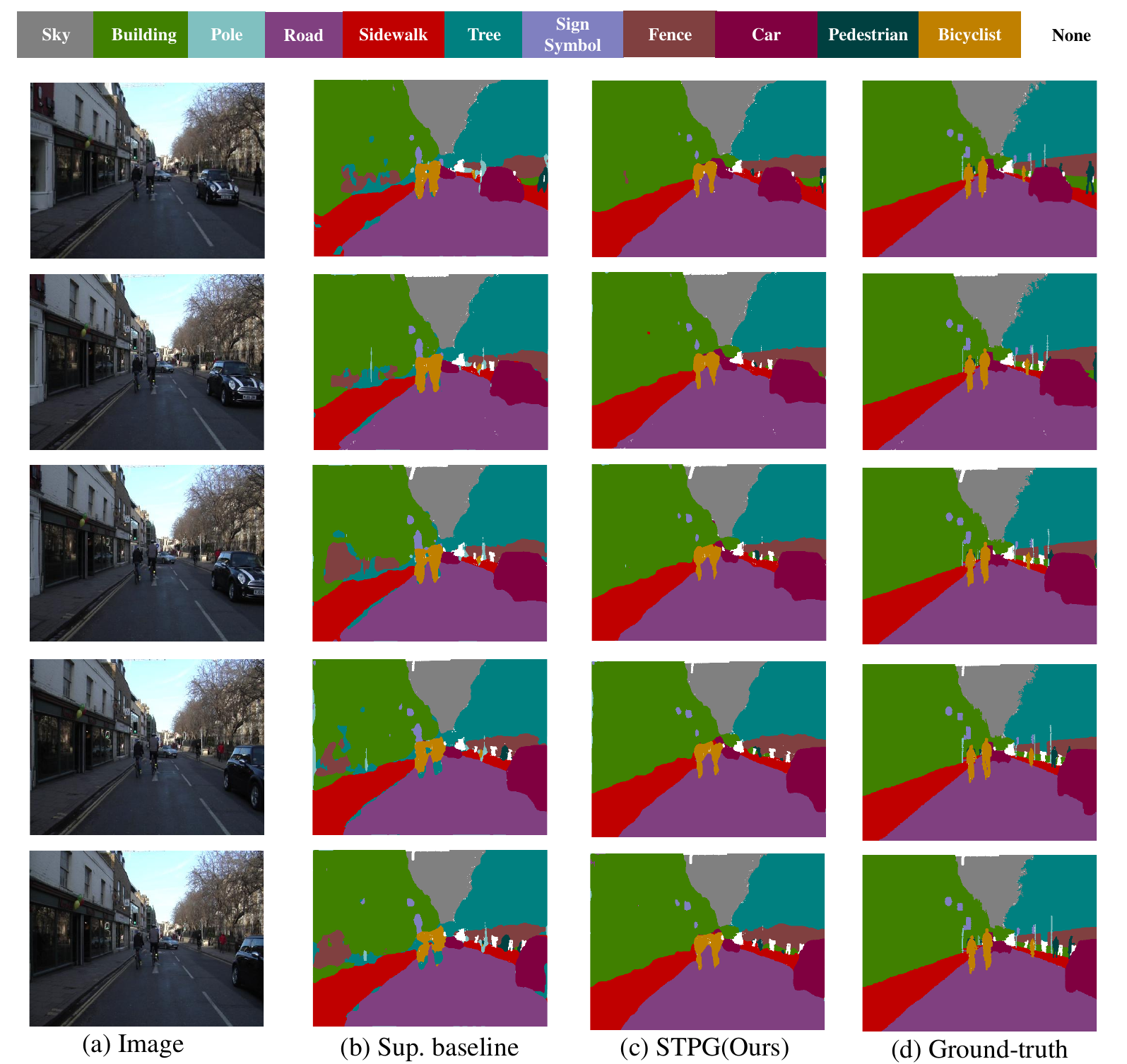}
	
\end{center}

\caption{Qualitative results of STPG framework using 1/8 labeled data from the CamVid training set. (a) Images from the validation set. (b) The results of training exclusively with labeled data. (c) The results achieved with STPG. (d) Ground-truths. All experiments are conducted using DeepLabV3+ with ResNet50.}
\label{vocimage}

\end{figure}

\subsection{Qualitative Results}
Fig. \ref{cityimage} and Fig. \ref{vocimage} present the qualitative results of STPG on the Cityscapes and CamVid datasets, respectively. It is evident that our method achieves substantial improvements over training using only labeled data, particularly in semi-supervised settings. Notably, significant performance gains are observed for minority classes, demonstrating the effectiveness of STPG in handling class imbalance and enhancing overall segmentation accuracy.

\section{Conclusion}
In this work, we propose a novel framework for semi-supervised semantic segmentation to enhance fine-grained road scene understanding in intelligent transportation systems, especially when labeled data is limited. The framework incorporates both specialized and general training modules, which help reduce model coupling and error accumulation while preserving richer semantic information for minority classes. Furthermore, the introduction of anchor contrastive learning alleviates the dominance of majority classes in the feature space. In intelligent transportation systems, accurate perception of all types of road users, such as pedestrians, cyclists, and traffic signs, is essential to ensure the safety and effectiveness of autonomous driving.
Our method significantly improves the recognition performance of minority classes without sacrificing the accuracy of majority classes. This balanced and dependable performance across all categories makes the proposed approach well-suited for real-world scenarios.

\bibliographystyle{IEEEtran}

\bibliography{cas-refs}

@article{yin2025semi,
  title={Semi-supervised semantic segmentation with multi-constraint consistency learning},
  author={Yin, Jianjian and Chen, Tao and Pei, Gensheng and Liu, Huafeng and Yao, Yazhou and Nie, Liqiang and Hua, Xiansheng},
  journal={IEEE Transactions on Multimedia},
  year={2025},
  publisher={IEEE}
}

@article{zhang2025noise,
  title={Noise-robust consistency regularization for semi-supervised semantic segmentation},
  author={Zhang, HaiKuan and Li, Haitao and Zhang, Xiufeng and Yang, Guanyu and Li, Atao and Du, Weisheng and Xue, Shanshan and Liu, Chi},
  journal={Neural Networks},
  volume={184},
  pages={107041},
  year={2025},
  publisher={Elsevier}
}

@inproceedings{lv2025scalematch,
  title={ScaleMatch: Multi-scale Consistency Enhancement for Semi-supervised Semantic Segmentation},
  author={Lv, Liang and Zhang, Lefei},
  booktitle={Proceedings of the AAAI Conference on Artificial Intelligence},
  volume={39},
  number={6},
  pages={5910--5918},
  year={2025}
}

@article{wang2025knowledge,
	title={Knowledge Generation and Distillation for Road Segmentation in Intelligent Transportation Systems},
	author={Wang, Jianyong and Gao, Mingliang and Zhai, Wenzhe and Rida, Imad and Zhu, Xianxun and Li, Qilei},
	journal={IEEE Transactions on Intelligent Transportation Systems},
	year={2025},
	publisher={IEEE}
}

@article{muhammad2022vision,
	title={Vision-based semantic segmentation in scene understanding for autonomous driving: Recent achievements, challenges, and outlooks},
	author={Muhammad, Khan and Hussain, Tanveer and Ullah, Hayat and Del Ser, Javier and Rezaei, Mahdi and Kumar, Neeraj and Hijji, Mohammad and Bellavista, Paolo and de Albuquerque, Victor Hugo C},
	journal={IEEE Transactions on Intelligent Transportation Systems},
	volume={23},
	number={12},
	pages={22694--22715},
	year={2022},
	publisher={IEEE}
}

@inproceedings{yeh2022decoupled,
	title={Decoupled contrastive learning},
	author={Yeh, Chun-Hsiao and Hong, Cheng-Yao and Hsu, Yen-Chi and Liu, Tyng-Luh and Chen, Yubei and LeCun, Yann},
	booktitle={European conference on computer vision},
	pages={668--684},
	year={2022},
	organization={Springer}
}

@inproceedings{cui2021parametric,
	title={Parametric contrastive learning},
	author={Cui, Jiequan and Zhong, Zhisheng and Liu, Shu and Yu, Bei and Jia, Jiaya},
	booktitle={Proceedings of the IEEE/CVF international conference on computer vision},
	pages={715--724},
	year={2021}
}

@article{wang2022contrastive,
	title={Contrastive learning with stronger augmentations},
	author={Wang, Xiao and Qi, Guo-Jun},
	journal={IEEE transactions on pattern analysis and machine intelligence},
	volume={45},
	number={5},
	pages={5549--5560},
	year={2022},
	publisher={IEEE}
}

@inproceedings{chen2022semi,
	title={Semi-supervised semantic segmentation via prototypical contrastive learning},
	author={Chen, Zenggui and Lian, Zhouhui},
	booktitle={Proceedings of the 30th ACM International Conference on Multimedia},
	pages={6696--6705},
	year={2022}
}

@inproceedings{yin2025uncertainty,
	title={Uncertainty-participation context consistency learning for semi-supervised semantic segmentation},
	author={Yin, Jianjian and Chen, Yi and Zheng, Zhichao and Zhou, Junsheng and Gu, Yanhui},
	booktitle={ICASSP 2025-2025 IEEE International Conference on Acoustics, Speech and Signal Processing (ICASSP)},
	pages={1--5},
	year={2025},
	organization={IEEE}
}

@inproceedings{sun2024corrmatch,
	title={Corrmatch: Label propagation via correlation matching for semi-supervised semantic segmentation},
	author={Sun, Boyuan and Yang, Yuqi and Zhang, Le and Cheng, Ming-Ming and Hou, Qibin},
	booktitle={Proceedings of the IEEE/CVF Conference on Computer Vision and Pattern Recognition},
	pages={3097--3107},
	year={2024}
}

@inproceedings{ma2023enhanced,
	title={Enhanced soft label for semi-supervised semantic segmentation},
	author={Ma, Jie and Wang, Chuan and Liu, Yang and Lin, Liang and Li, Guanbin},
	booktitle={Proceedings of the IEEE/CVF International Conference on Computer Vision},
	pages={1185--1195},
	year={2023}
}

@inproceedings{yang2023revisiting,
	title={Revisiting weak-to-strong consistency in semi-supervised semantic segmentation},
	author={Yang, Lihe and Qi, Lei and Feng, Litong and Zhang, Wayne and Shi, Yinghuan},
	booktitle={Proceedings of the IEEE/CVF conference on computer vision and pattern recognition},
	pages={7236--7246},
	year={2023}
}

@inproceedings{brostow2008segmentation,
	title={Segmentation and recognition using structure from motion point clouds},
	author={Brostow, Gabriel J and Shotton, Jamie and Fauqueur, Julien and Cipolla, Roberto},
	booktitle={Computer vision--ECCV 2008: 10th European conference on computer vision, marseille, France, October 12-18, 2008, proceedings, part i 10},
	pages={44--57},
	year={2008},
	organization={Springer}
}

@inproceedings{wang2023space,
	title={Space engage: Collaborative space supervision for contrastive-based semi-supervised semantic segmentation},
	author={Wang, Changqi and Xie, Haoyu and Yuan, Yuhui and Fu, Chong and Yue, Xiangyu},
	booktitle={Proceedings of the IEEE/CVF International Conference on Computer Vision},
	pages={931--942},
	year={2023}
}

@inproceedings{fang2023locating,
	title={Locating noise is halfway denoising for semi-supervised segmentation},
	author={Fang, Yan and Zhu, Feng and Cheng, Bowen and Liu, Luoqi and Zhao, Yao and Wei, Yunchao},
	booktitle={Proceedings of the IEEE/CVF International Conference on Computer Vision},
	pages={16612--16622},
	year={2023}
}

@inproceedings{huang2023semicvt,
	title={Semicvt: Semi-supervised convolutional vision transformer for semantic segmentation},
	author={Huang, Huimin and Xie, Shiao and Lin, Lanfen and Tong, Ruofeng and Chen, Yen-Wei and Li, Yuexiang and Wang, Hong and Huang, Yawen and Zheng, Yefeng},
	booktitle={Proceedings of the IEEE/CVF Conference on Computer Vision and Pattern Recognition},
	pages={11340--11349},
	year={2023}
}

@article{hu2024multi,
	title={Multi-Perspective Pseudo-Label Generation and Confidence-Weighted Training for Semi-Supervised Semantic Segmentation},
	author={Hu, Kai and Chen, Xiaobo and Chen, Zhineng and Zhang, Yuan and Gao, Xieping},
	journal={IEEE Transactions on Multimedia},
	year={2024},
	publisher={IEEE}
}

@article{hong2024multi,
	title={A multi-view consistency framework with semi-supervised domain adaptation},
	author={Hong, Yuting and Dong, Li and Qiu, Xiaojie and Xiao, Hui and Yao, Baochen and Zheng, Siming and Peng, Chengbin},
	journal={Engineering Applications of Artificial Intelligence},
	volume={136},
	pages={108886},
	year={2024},
	publisher={Elsevier}
}

@inproceedings{li2020overcoming,
	title={Overcoming classifier imbalance for long-tail object detection with balanced group softmax},
	author={Li, Yu and Wang, Tao and Kang, Bingyi and Tang, Sheng and Wang, Chunfeng and Li, Jintao and Feng, Jiashi},
	booktitle={Proceedings of the IEEE/CVF conference on computer vision and pattern recognition},
	pages={10991--11000},
	year={2020}
}

@article{cao2019learning,
	title={Learning imbalanced datasets with label-distribution-aware margin loss},
	author={Cao, Kaidi and Wei, Colin and Gaidon, Adrien and Arechiga, Nikos and Ma, Tengyu},
	journal={Advances in neural information processing systems},
	volume={32},
	year={2019}
}

@article{liao2024pda,
	title={PDA: Progressive Domain Adaptation for Semantic Segmentation},
	author={Liao, Muxin and Tian, Shishun and Zhang, Yuhang and Hua, Guoguang and Zou, Wenbin and Li, Xia},
	journal={Knowledge-Based Systems},
	volume={284},
	pages={111179},
	year={2024},
	publisher={Elsevier}
}

@article{chang2022fully,
	title={Fully used reliable data and attention consistency for semi-supervised learning},
	author={Chang, Jui-Hung and Weng, Hsiu-Chen},
	journal={Knowledge-Based Systems},
	volume={249},
	pages={108837},
	year={2022},
	publisher={Elsevier}
}

@inproceedings{feng2024bacon,
	title={BaCon: Boosting Imbalanced Semi-supervised Learning via Balanced Feature-Level Contrastive Learning},
	author={Feng, Qianhan and Xie, Lujing and Fang, Shijie and Lin, Tong},
	booktitle={Proceedings of the AAAI Conference on Artificial Intelligence},
	volume={38},
	number={11},
	pages={11970--11978},
	year={2024}
}

@article{chen2022debiased,
	title={Debiased self-training for semi-supervised learning},
	author={Chen, Baixu and Jiang, Junguang and Wang, Ximei and Wan, Pengfei and Wang, Jianmin and Long, Mingsheng},
	journal={Advances in Neural Information Processing Systems},
	volume={35},
	pages={32424--32437},
	year={2022}
}

@inproceedings{qiao2023fuzzy,
	title={Fuzzy positive learning for semi-supervised semantic segmentation},
	author={Qiao, Pengchong and Wei, Zhidan and Wang, Yu and Wang, Zhennan and Song, Guoli and Xu, Fan and Ji, Xiangyang and Liu, Chang and Chen, Jie},
	booktitle={Proceedings of the IEEE/CVF Conference on Computer Vision and Pattern Recognition},
	pages={15465--15474},
	year={2023}
}

@inproceedings{yun2019cutmix,
	title={Cutmix: Regularization strategy to train strong classifiers with localizable features},
	author={Yun, Sangdoo and Han, Dongyoon and Oh, Seong Joon and Chun, Sanghyuk and Choe, Junsuk and Yoo, Youngjoon},
	booktitle={Proceedings of the IEEE/CVF international conference on computer vision},
	pages={6023--6032},
	year={2019}
}

@article{chen2023adversarial,
	title={Adversarial learning of object-aware activation map for weakly-supervised semantic segmentation},
	author={Chen, Junliang and Lu, Weizeng and Li, Yuexiang and Shen, Linlin and Duan, Jinming},
	journal={IEEE Transactions on Circuits and Systems for Video Technology},
	year={2023},
	publisher={IEEE}
}

@article{duan2022mutexmatch,
	title={Mutexmatch: semi-supervised learning with mutex-based consistency regularization},
	author={Duan, Yue and Zhao, Zhen and Qi, Lei and Wang, Lei and Zhou, Luping and Shi, Yinghuan and Gao, Yang},
	journal={IEEE Transactions on Neural Networks and Learning Systems},
	year={2022},
	publisher={IEEE}
}

@inproceedings{kim2022conmatch,
	title={Conmatch: Semi-supervised learning with confidence-guided consistency regularization},
	author={Kim, Jiwon and Min, Youngjo and Kim, Daehwan and Lee, Gyuseong and Seo, Junyoung and Ryoo, Kwangrok and Kim, Seungryong},
	booktitle={European Conference on Computer Vision},
	pages={674--690},
	year={2022},
	organization={Springer}
}

@inproceedings{zhou2023hypermatch,
	title={HyperMatch: Noise-tolerant semi-supervised learning via relaxed contrastive constraint},
	author={Zhou, Beitong and Lu, Jing and Liu, Kerui and Xu, Yunlu and Cheng, Zhanzhan and Niu, Yi},
	booktitle={Proceedings of the IEEE/CVF Conference on Computer Vision and Pattern Recognition},
	pages={24017--24026},
	year={2023}
}

@inproceedings{zheng2021rethinking,
	title={Rethinking semantic segmentation from a sequence-to-sequence perspective with transformers},
	author={Zheng, Sixiao and Lu, Jiachen and Zhao, Hengshuang and Zhu, Xiatian and Luo, Zekun and Wang, Yabiao and Fu, Yanwei and Feng, Jianfeng and Xiang, Tao and Torr, Philip HS and others},
	booktitle={Proceedings of the IEEE/CVF conference on computer vision and pattern recognition},
	pages={6881--6890},
	year={2021}
}

@article{guo2022segnext,
	title={Segnext: Rethinking convolutional attention design for semantic segmentation},
	author={Guo, Meng-Hao and Lu, Cheng-Ze and Hou, Qibin and Liu, Zhengning and Cheng, Ming-Ming and Hu, Shi-Min},
	journal={Advances in Neural Information Processing Systems},
	volume={35},
	pages={1140--1156},
	year={2022}
}

@inproceedings{cheng2022masked,
	title={Masked-attention mask transformer for universal image segmentation},
	author={Cheng, Bowen and Misra, Ishan and Schwing, Alexander G and Kirillov, Alexander and Girdhar, Rohit},
	booktitle={Proceedings of the IEEE/CVF conference on computer vision and pattern recognition},
	pages={1290--1299},
	year={2022}
}

@article{vaswani2017attention,
	title={Attention is all you need},
	author={Vaswani, Ashish and Shazeer, Noam and Parmar, Niki and Uszkoreit, Jakob and Jones, Llion and Gomez, Aidan N and Kaiser, {\L}ukasz and Polosukhin, Illia},
	journal={Advances in neural information processing systems},
	volume={30},
	year={2017}
}

@article{lecun2015deep,
	title={Deep learning},
	author={LeCun, Yann and Bengio, Yoshua and Hinton, Geoffrey},
	journal={nature},
	volume={521},
	number={7553},
	pages={436--444},
	year={2015},
	publisher={Nature Publishing Group UK London}
}

@inproceedings{xie2023boosting,
	title={Boosting semi-supervised semantic segmentation with probabilistic representations},
	author={Xie, Haoyu and Wang, Changqi and Zheng, Mingkai and Dong, Minjing and You, Shan and Fu, Chong and Xu, Chang},
	booktitle={Proceedings of the AAAI Conference on Artificial Intelligence},
	volume={37},
	number={3},
	pages={2938--2946},
	year={2023}
}

@article{yin2023semi,
	title={Semi-supervised semantic segmentation with multi-reliability and multi-level feature augmentation},
	author={Yin, Jianjian and Zheng, Zhichao and Pan, Yulu and Gu, Yanhui and Chen, Yi},
	journal={Expert Systems with Applications},
	volume={233},
	pages={120973},
	year={2023},
	publisher={Elsevier}
}

@article{hou2024view,
	title={View-coherent correlation consistency for semi-supervised semantic segmentation},
	author={Hou, Yunzhong and Gould, Stephen and Zheng, Liang},
	journal={Pattern Recognition},
	volume={147},
	pages={110089},
	year={2024},
	publisher={Elsevier}
}

@inproceedings{hu2021simple,
	title={Simple: Similar pseudo label exploitation for semi-supervised classification},
	author={Hu, Zijian and Yang, Zhengyu and Hu, Xuefeng and Nevatia, Ram},
	booktitle={Proceedings of the IEEE/CVF Conference on Computer Vision and Pattern Recognition},
	pages={15099--15108},
	year={2021}
}

@article{xie2021segformer,
	title={SegFormer: Simple and efficient design for semantic segmentation with transformers},
	author={Xie, Enze and Wang, Wenhai and Yu, Zhiding and Anandkumar, Anima and Alvarez, Jose M and Luo, Ping},
	journal={Advances in Neural Information Processing Systems},
	volume={34},
	pages={12077--12090},
	year={2021}
}

@article{ji2020encoder,
	title={Encoder-decoder with cascaded CRFs for semantic segmentation},
	author={Ji, Jian and Shi, Rui and Li, Sitong and Chen, Peng and Miao, Qiguang},
	journal={IEEE Transactions on Circuits and Systems for Video Technology},
	volume={31},
	number={5},
	pages={1926--1938},
	year={2020},
	publisher={IEEE}
}

@article{zhao2022source,
	title={Source-free open compound domain adaptation in semantic segmentation},
	author={Zhao, Yuyang and Zhong, Zhun and Luo, Zhiming and Lee, Gim Hee and Sebe, Nicu},
	journal={IEEE Transactions on Circuits and Systems for Video Technology},
	volume={32},
	number={10},
	pages={7019--7032},
	year={2022},
	publisher={IEEE}
}

@article{meng2019weakly,
	title={Weakly supervised semantic segmentation by a class-level multiple group cosegmentation and foreground fusion strategy},
	author={Meng, Fanman and Luo, Kunming and Li, Hongliang and Wu, Qingbo and Xu, Xiaolong},
	journal={IEEE Transactions on Circuits and Systems for Video Technology},
	volume={30},
	number={12},
	pages={4823--4836},
	year={2019},
	publisher={IEEE}
}

@article{zhou2022context,
	title={Context-aware mixup for domain adaptive semantic segmentation},
	author={Zhou, Qianyu and Feng, Zhengyang and Gu, Qiqi and Pang, Jiangmiao and Cheng, Guangliang and Lu, Xuequan and Shi, Jianping and Ma, Lizhuang},
	journal={IEEE Transactions on Circuits and Systems for Video Technology},
	year={2022},
	publisher={IEEE}
}

@article{cao2022adversarial,
	title={Adversarial dual-student with differentiable spatial warping for semi-supervised semantic segmentation},
	author={Cao, Cong and Lin, Tianwei and He, Dongliang and Li, Fu and Yue, Huanjing and Yang, Jingyu and Ding, Errui},
	journal={IEEE Transactions on Circuits and Systems for Video Technology},
	year={2022},
	publisher={IEEE}
}

@article{bachman2014learning,
	title={Learning with pseudo-ensembles},
	author={Bachman, Philip and Alsharif, Ouais and Precup, Doina},
	journal={Advances in neural information processing systems},
	volume={27},
	year={2014}
}

@article{jonker1986improving,
	title={Improving the Hungarian assignment algorithm},
	author={Jonker, Roy and Volgenant, Ton},
	journal={Operations Research Letters},
	volume={5},
	number={4},
	pages={171--175},
	year={1986},
	publisher={Elsevier}
}

@inproceedings{hariharan2011semantic,
	title={Semantic contours from inverse detectors},
	author={Hariharan, Bharath and Arbel{\'a}ez, Pablo and Bourdev, Lubomir and Maji, Subhransu and Malik, Jitendra},
	booktitle={2011 international conference on computer vision},
	pages={991--998},
	year={2011},
	organization={IEEE}
}

@article{everingham2010pascal,
	title={The pascal visual object classes (voc) challenge},
	author={Everingham, Mark and Van Gool, Luc and Williams, Christopher KI and Winn, John and Zisserman, Andrew},
	journal={International journal of computer vision},
	volume={88},
	pages={303--338},
	year={2010},
	publisher={Springer}
}

@article{fan2023conservative,
  title={Conservative-progressive collaborative learning for semi-supervised semantic segmentation},
  author={Fan, Siqi and Zhu, Fenghua and Feng, Zunlei and Lv, Yisheng and Song, Mingli and Wang, Fei-Yue},
  journal={IEEE Transactions on Image Processing},
  year={2023},
  publisher={IEEE}
}

@article{xiao2022semi,
  title={Semi-supervised semantic segmentation with cross teacher training},
  author={Xiao, Hui and Li, Dong and Xu, Hao and Fu, Shuibo and Yan, Diqun and Song, Kangkang and Peng, Chengbin},
  journal={Neurocomputing},
  volume={508},
  pages={36--46},
  year={2022},
  publisher={Elsevier}
}

@inproceedings{yang2022st++,
  title={St++: Make self-training work better for semi-supervised semantic segmentation},
  author={Yang, Lihe and Zhuo, Wei and Qi, Lei and Shi, Yinghuan and Gao, Yang},
  booktitle={Proceedings of the IEEE/CVF Conference on Computer Vision and Pattern Recognition},
  pages={4268--4277},
  year={2022}
}

@inproceedings{kong2023pruning,
  title={Pruning-Guided Curriculum Learning for Semi-Supervised Semantic Segmentation},
  author={Kong, Heejo and Lee, Gun-Hee and Kim, Suneung and Lee, Seong-Whan},
  booktitle={Proceedings of the IEEE/CVF Winter Conference on Applications of Computer Vision},
  pages={5914--5923},
  year={2023}
}

@inproceedings{kwon2022semi,
  title={Semi-supervised semantic segmentation with error localization network},
  author={Kwon, Donghyeon and Kwak, Suha},
  booktitle={Proceedings of the IEEE/CVF Conference on Computer Vision and Pattern Recognition},
  pages={9957--9967},
  year={2022}
}

@inproceedings{zhong2021pixel,
  title={Pixel contrastive-consistent semi-supervised semantic segmentation},
  author={Zhong, Yuanyi and Yuan, Bodi and Wu, Hong and Yuan, Zhiqiang and Peng, Jian and Wang, Yu-Xiong},
  booktitle={Proceedings of the IEEE/CVF International Conference on Computer Vision},
  pages={7273--7282},
  year={2021}
}

@inproceedings{lai2021semi,
  title={Semi-supervised semantic segmentation with directional context-aware consistency},
  author={Lai, Xin and Tian, Zhuotao and Jiang, Li and Liu, Shu and Zhao, Hengshuang and Wang, Liwei and Jia, Jiaya},
  booktitle={Proceedings of the IEEE/CVF Conference on Computer Vision and Pattern Recognition},
  pages={1205--1214},
  year={2021}
}

@inproceedings{wang2020understanding,
  title={Understanding contrastive representation learning through alignment and uniformity on the hypersphere},
  author={Wang, Tongzhou and Isola, Phillip},
  booktitle={International Conference on Machine Learning},
  pages={9929--9939},
  year={2020},
  organization={PMLR}
}

@inproceedings{lee2013pseudo,
	title={Pseudo-label: The simple and efficient semi-supervised learning method for deep neural networks},
	author={Lee, Dong-Hyun and others},
	booktitle={Workshop on challenges in representation learning, ICML},
	volume={3},
	number={2},
	pages={896},
	year={2013}
}

@article{hu2021semi,
	title={Semi-supervised semantic segmentation via adaptive equalization learning},
	author={Hu, Hanzhe and Wei, Fangyun and Hu, Han and Ye, Qiwei and Cui, Jinshi and Wang, Liwei},
	journal={Advances in Neural Information Processing Systems},
	volume={34},
	pages={22106--22118},
	year={2021}
}

@inproceedings{wang2022semi,
	title={Semi-Supervised Semantic Segmentation Using Unreliable Pseudo-Labels},
	author={Wang, Yuchao and Wang, Haochen and Shen, Yujun and Fei, Jingjing and Li, Wei and Jin, Guoqiang and Wu, Liwei and Zhao, Rui and Le, Xinyi},
	booktitle={Proceedings of the IEEE/CVF Conference on Computer Vision and Pattern Recognition},
	pages={4248--4257},
	year={2022}
}

@article{sohn2020fixmatch,
	title={Fixmatch: Simplifying semi-supervised learning with consistency and confidence},
	author={Sohn, Kihyuk and Berthelot, David and Carlini, Nicholas and Zhang, Zizhao and Zhang, Han and Raffel, Colin A and Cubuk, Ekin Dogus and Kurakin, Alexey and Li, Chun-Liang},
	journal={Advances in neural information processing systems},
	volume={33},
	pages={596--608},
	year={2020}
}

@article{feng2022dmt,
	title={Dmt: Dynamic mutual training for semi-supervised learning},
	author={Feng, Zhengyang and Zhou, Qianyu and Gu, Qiqi and Tan, Xin and Cheng, Guangliang and Lu, Xuequan and Shi, Jianping and Ma, Lizhuang},
	journal={Pattern Recognition},
	pages={108777},
	year={2022},
	publisher={Elsevier}
}

@inproceedings{liu2022perturbed,
	title={Perturbed and strict mean teachers for semi-supervised semantic segmentation},
	author={Liu, Yuyuan and Tian, Yu and Chen, Yuanhong and Liu, Fengbei and Belagiannis, Vasileios and Carneiro, Gustavo},
	booktitle={Proceedings of the IEEE/CVF Conference on Computer Vision and Pattern Recognition},
	pages={4258--4267},
	year={2022}
}

@inproceedings{wei2021crest,
	title={Crest: A class-rebalancing self-training framework for imbalanced semi-supervised learning},
	author={Wei, Chen and Sohn, Kihyuk and Mellina, Clayton and Yuille, Alan and Yang, Fan},
	booktitle={Proceedings of the IEEE/CVF Conference on Computer Vision and Pattern Recognition},
	pages={10857--10866},
	year={2021}
}

@inproceedings{ke2020guided,
	title={Guided collaborative training for pixel-wise semi-supervised learning},
	author={Ke, Zhanghan and Qiu, Di and Li, Kaican and Yan, Qiong and Lau, Rynson WH},
	booktitle={Proceedings of the European conference on computer vision},
	pages={429--445},
	year={2020},
}

@article{tarvainen2017mean,
	title={Mean teachers are better role models: Weight-averaged consistency targets improve semi-supervised deep learning results},
	author={Tarvainen, Antti and Valpola, Harri},
	journal={arXiv preprint arXiv:1703.01780},
	year={2017}
}

@inproceedings{french2019semi,
	title={Semi-supervised semantic segmentation needs strong, varied perturbations},
	author={French, Geoffrey and Laine, Samuli and Aila, Timo and Mackiewicz, Michal and Finlayson, Graham},
	booktitle={British Machine Vision Conference},
	number={31},
	year={2020}
}

@inproceedings{ke2019dual,
	title={Dual student: Breaking the limits of the teacher in semi-supervised learning},
	author={Ke, Zhanghan and Wang, Daoye and Yan, Qiong and Ren, Jimmy and Lau, Rynson WH},
	booktitle={Proceedings of the IEEE/CVF International Conference on Computer Vision},
	pages={6728--6736},
	year={2019}
}

@inproceedings{cordts2016cityscapes,
	title={The cityscapes dataset for semantic urban scene understanding},
	author={Cordts, Marius and Omran, Mohamed and Ramos, Sebastian and Rehfeld, Timo and Enzweiler, Markus and Benenson, Rodrigo and Franke, Uwe and Roth, Stefan and Schiele, Bernt},
	booktitle={Proceedings of the IEEE conference on computer vision and pattern recognition},
	pages={3213--3223},
	year={2016}
}

@inproceedings{long2015fully,
	title={Fully convolutional networks for semantic segmentation},
	author={Long, Jonathan and Shelhamer, Evan and Darrell, Trevor},
	booktitle={Proceedings of the IEEE conference on computer vision and pattern recognition},
	pages={3431--3440},
	year={2015}
}

@inproceedings{arazo2020pseudo,
	title={Pseudo-labeling and confirmation bias in deep semi-supervised learning},
	author={Arazo, Eric and Ortego, Diego and Albert, Paul and O’Connor, Noel E and McGuinness, Kevin},
	booktitle={International Joint Conference on Neural Networks},
	pages={1--8},
	year={2020},
	organization={IEEE}
}

@inproceedings{chen2018encoder,
	title={Encoder-decoder with atrous separable convolution for semantic image segmentation},
	author={Chen, Liang-Chieh and Zhu, Yukun and Papandreou, George and Schroff, Florian and Adam, Hartwig},
	booktitle={Proceedings of the European conference on computer vision},
	pages={801--818},
	year={2018}
}

@inproceedings{alonso2021semi,
	title={Semi-supervised semantic segmentation with pixel-level contrastive learning from a class-wise memory bank},
	author={Alonso, Inigo and Sabater, Alberto and Ferstl, David and Montesano, Luis and Murillo, Ana C},
	booktitle={Proceedings of the IEEE/CVF International Conference on Computer Vision},
	pages={8219--8228},
	year={2021}
}

@inproceedings{ouali2020semi,
	title={Semi-supervised semantic segmentation with cross-consistency training},
	author={Ouali, Yassine and Hudelot, C{\'e}line and Tami, Myriam},
	booktitle={Proceedings of the IEEE/CVF Conference on Computer Vision and Pattern Recognition},
	pages={12674--12684},
	year={2020}
}

@inproceedings{chen2021semi,
	title={Semi-Supervised Semantic Segmentation with Cross Pseudo Supervision},
	author={Chen, Xiaokang and Yuan, Yuhui and Zeng, Gang and Wang, Jingdong},
	booktitle={Proceedings of the IEEE/CVF Conference on Computer Vision and Pattern Recognition},
	pages={2613--2622},
	year={2021}
}

@article{zou2020pseudoseg,
	title={PseudoSeg: Designing Pseudo Labels for Semantic Segmentation},
	author={Zou, Yuliang and Zhang, Zizhao and Zhang, Han and Li, Chun-Liang and Bian, Xiao and Huang, Jia-Bin and Pfister, Tomas},
	journal={International Conference on Learning Representations},
	year={2021}
}

@inproceedings{deng2009imagenet,
	title={Imagenet: A large-scale hierarchical image database},
	author={Deng, Jia and Dong, Wei and Socher, Richard and Li, Li-Jia and Li, Kai and Fei-Fei, Li},
	booktitle={2009 IEEE conference on computer vision and pattern recognition},
	pages={248--255},
	year={2009},
	organization={Ieee}
}


\begin{IEEEbiography}[{\includegraphics[width=1in,height=1in,clip,keepaspectratio]{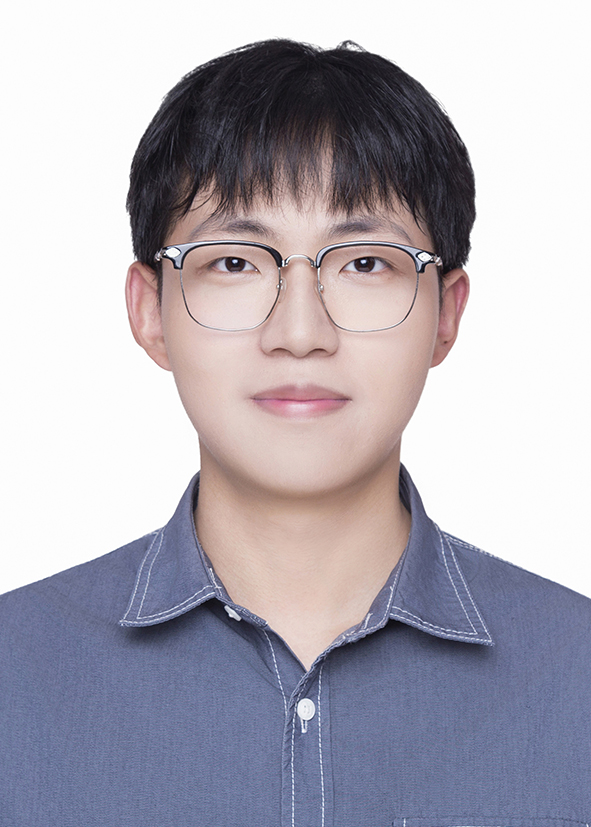}}]{Yuting Hong}
	is a graduate student at the Faculty of Electrical Engineering and Computer Science, Ningbo University. His research interests include semi-supervised learning, domain adaptation, and semantic segmentation.
\end{IEEEbiography}

\vspace{-50pt}
\begin{IEEEbiography}[{\includegraphics[width=1in,height=1in,clip,keepaspectratio]{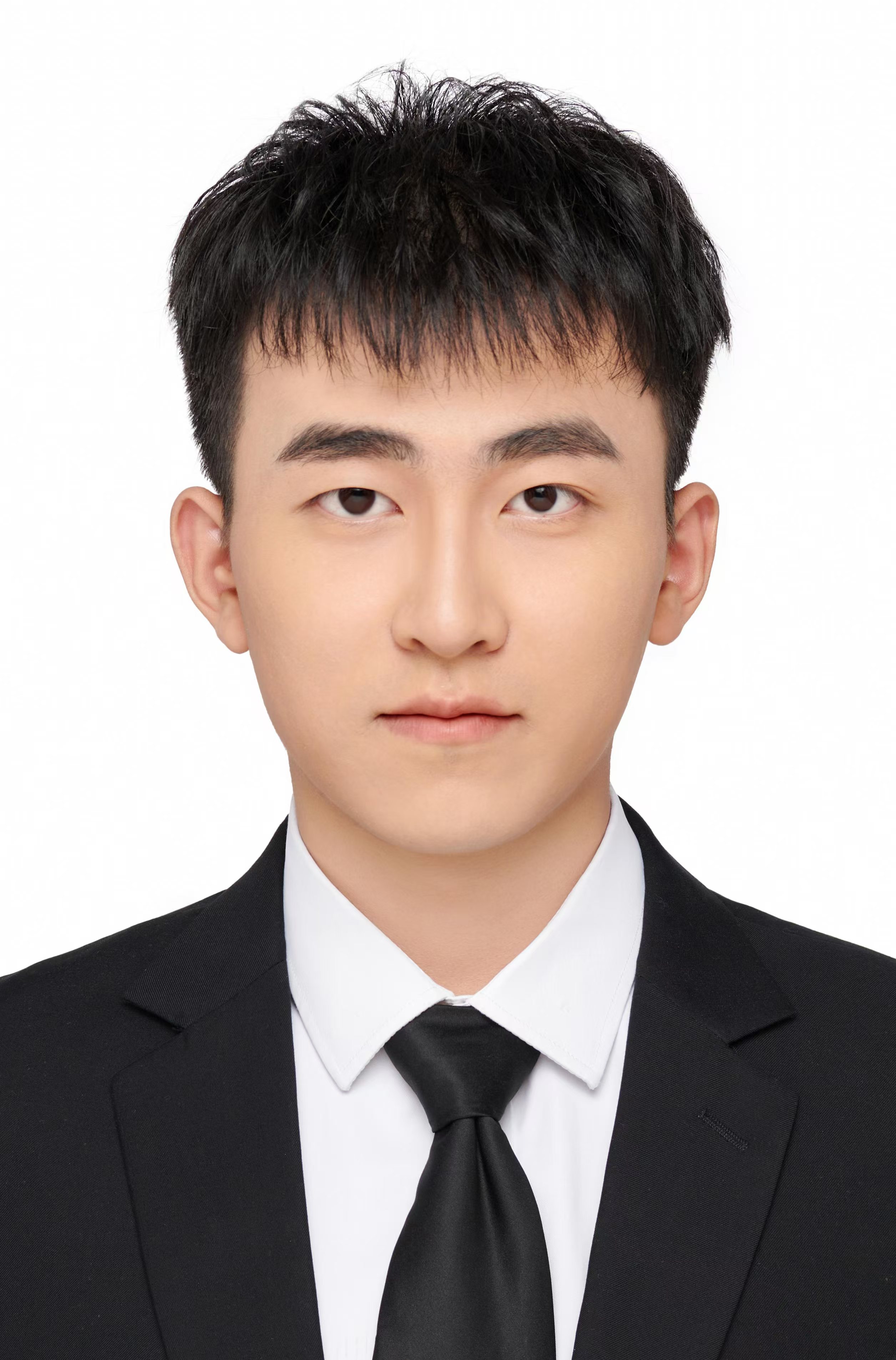}}]{Yongkang Wu}
	is a graduate student at the College of Information Science and Engineering, Ningbo University. His research focuses on semi-supervised learning, object detection, and table structure recognition.
\end{IEEEbiography}

\vspace{-50pt}
\begin{IEEEbiography}[{\includegraphics[width=1in,height=1in,clip,keepaspectratio]{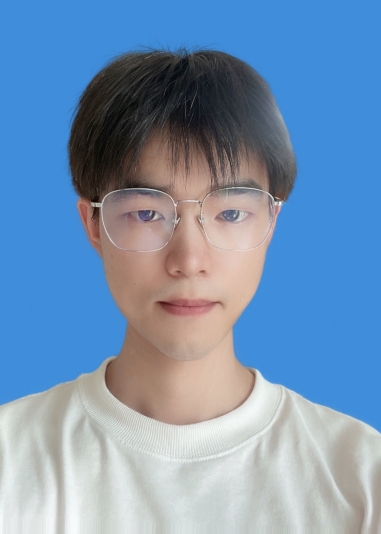}}]{Hui Xiao}
is a graduate student at the Faculty of Electrical Engineering and Computer Science, Ningbo University. His research interests include semi-supervised learning, semi-supervised semantic segmentation, and semantic segmentation.
\end{IEEEbiography}
\vspace{-50pt}
\begin{IEEEbiography}[{\includegraphics[width=1in,height=1in,clip,keepaspectratio]{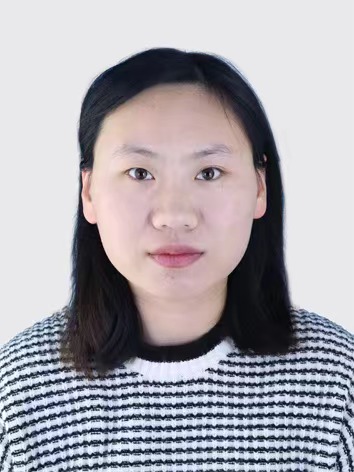}}]{Huazheng Hao}
 is a graduate student at the Faculty of Electrical Engineering and Computer Science, Ningbo University. Her research interests include semi-supervised learning, semi-supervised semantic segmentation, and semantic segmentation.
\end{IEEEbiography}
\vspace{-50pt}
\begin{IEEEbiography}[{\includegraphics[width=1in,height=1in,clip,keepaspectratio]{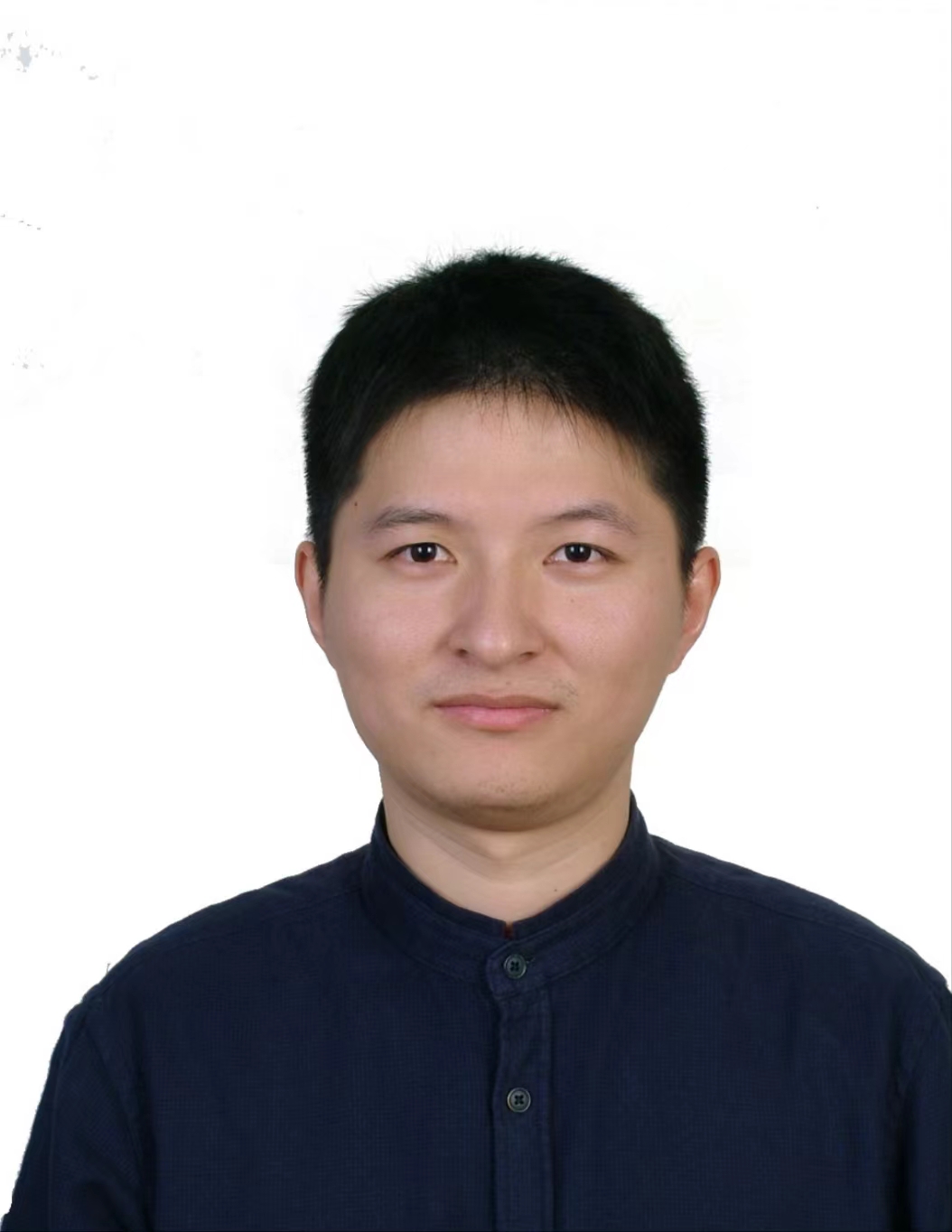}}]{Xiaojie Qiu}
received the M.Eng. degree from Zhejiang University of Technology in 2014. He is currently managing the automation system group in Cowain and is responsible for developing robotic applications. His research interests include automation systems and  computer vision.
\end{IEEEbiography}
\vspace{-50pt}
\begin{IEEEbiography}[{\includegraphics[width=1in,height=1in,clip,keepaspectratio]{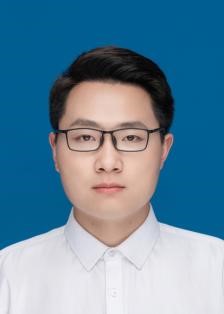}}]{Baochen Yao}
	received the M.S. degree in signal and Information processing from Anhui University, Hefei, China. He is currently pursuing the Ph.D. degree with the Faculty of Electrical Engineering and Computer Science, Ningbo University, China. His current research interests include computer vision, weakly supervised learning, and 3D point cloud processing.
\end{IEEEbiography}
\vspace{-50pt}
\begin{IEEEbiography}[{\includegraphics[width=1in,height=1in,clip,keepaspectratio]{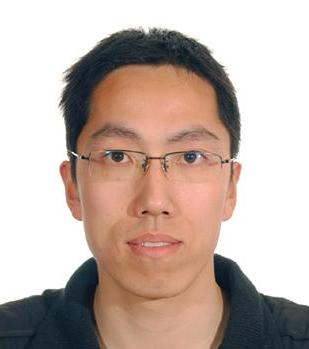}}]{Chengbin Peng (Member, IEEE)}
	received the B.E. and M.S. degrees in computer science from Zhejiang University, China. He received a Ph.D. degree in computer science from King Abdullah University of Science and Technology, KSA, in 2015. He is currently an Associate Professor with the Faculty of Electrical Engineering and Computer Science, Ningbo University. His research interests include computer vision,  complex network analysis, and semi-supervised learning.
\end{IEEEbiography}

%

\end{document}